\documentclass[10pt,twocolumn,letterpaper]{article}

\usepackage{cvpr}      % To produce the REVIEW version
\usepackage{algorithm}
\usepackage{url}
\usepackage{algpseudocode}
\newcommand{\sysname}{AdaCluster\xspace}

\definecolor{darkgreen}{rgb}{0.078,0.667,0.016}

\newcommand{\clustername}{AdaCluster\xspace}

\newcommand{\para}[1]{\noindent \textbf{#1 }}
\definecolor{cvprblue}{rgb}{0.21,0.49,0.74}
\usepackage[pagebackref,breaklinks,colorlinks,allcolors=cvprblue]{hyperref}
\usepackage{dblfloatfix}
\usepackage{caption}
\usepackage{cuted}
\usepackage{tabularx}
\usepackage[table]{xcolor}
\usepackage{comment}
\def\paperID{} % *** Enter the Paper ID here
\def\confName{CVPR}
\def\confYear{2026}

\title{\sysname: Adaptive Query-Key Clustering for Sparse Attention in Video Generation}
\author{
    Haoyue Tan\textsuperscript{1,2}\thanks{Equal contribution.}, 
    Shengnan Wang\textsuperscript{3}\footnotemark[1], Yulin 
    Qiao\textsuperscript{4}, Juncheng Zhang\textsuperscript{1,5}, Youhui Bai\textsuperscript{1}\thanks{Corresponding author, youhuibai@ustc.edu.cn.}, \\
     Ping Gong\textsuperscript{1}, Zewen Jin\textsuperscript{1}, Cheng Li\textsuperscript{1,2} \\
    \textsuperscript{1}University of Science and Technology of China \\
    \textsuperscript{2}Institute of Artificial Intelligence, Hefei Comprehensive National Science Center \\
    \textsuperscript{3}Independent Researcher\\
    \textsuperscript{4}University of Macau\\
    \textsuperscript{5}The Chinese University of Hong Kong
}
\begin{document}

\maketitle

% \begin{strip}

%     \centering
%     \includegraphics[width=0.8\textwidth]{upspeed.pdf}
%     \captionof{figure}{Performance comparison between Full Attention and \clustername methods on HunyuanVideo, with videos generated on a single A40. \clustername reduces inference latency from 1639s to 973s with a PSNR of 30.58dB, significantly improving inference speed while maintaining visual quality. }
%     \label{fig:placeholder}

% \end{strip}

\begin{abstract}

Video diffusion transformers (DiTs) suffer from prohibitive inference latency due to quadratic attention complexity. Existing sparse attention methods either overlook semantic similarity, or fail to adapt to heterogeneous token distributions across layers, leading to model performance degradation. We propose \sysname, a training‑free adaptive clustering framework that accelerates the generation of DiTs while preserving accuracy. \sysname applies an angle-similarity preserving clustering method to query vectors for higher compression, and designs a euclidean-similarity preserving clustering method for keys, covering cluster number assignment, threshold-wise adaptive clustering, and efficient critical cluster selection. Experiments on CogVideoX‑2B, HunyuanVideo, and Wan‑2.1 via one A40 GPU demonstrate up to 1.67$\times$-4.31$\times$ speedup with negligible quality degradation. 
% ; notably, inference time on HunyuanVideo (180K tokens) is reduced from 1639s to 973s with a PSNR of 30.58dB, substantially outperforming prior sparse attention methods.
% }

\noindent
Code: \href{https://github.com/USTC-MLSys-Team/Adacluster}{%
   {https://github.com/USTC-MLSys-Team/Adacluster}%
}

\end{abstract}

\section{Introduction}
\label{sec:intro}

%jc Video diffusion models have made great progress in generating high-resolution and temporally consistent videos. They are now widely used in animation generation and physical scene simulation. However, these models still need very high computation. The main cost comes from the attention mechanism, which has quadratic complexity. For example, in \textbf{CogVideoX-2B}, the context length can reach 70K--100K, and attention takes more than 75\% of the total inference time. In \textbf{HunyuanVideo}, when the context grows to 180K, attention cost is over 80\%. As video resolution and frame number increase, this problem becomes even more serious, making attention the main obstacle to efficient video generation.

% Diffusion models have become increasingly popular for video generation and physical scene simulation~\cite{Peebles_2023_ICCV,blattmann2023align,croitoru2023diffusion}. 
Diffusion transformer models (DiTs) have emerged as a dominant paradigm for video generation and physical scene simulation, demonstrating strong scalability and high-fidelity synthesis capabilities~\cite{Peebles_2023_ICCV,blattmann2023align,croitoru2023diffusion}.
However, generation latency remains a major bottleneck due to inherently high computational complexity~\cite{wang2024recipe,wang2023modelscope,chen2024videocrafter2}. 
% The primary contributor to this cost is the attention modules, whose complexity grow quadratically with the input sequence length~\cite{vaswani2017attention}. 
\begin{figure}
    \centering
    \includegraphics[width=1\linewidth]{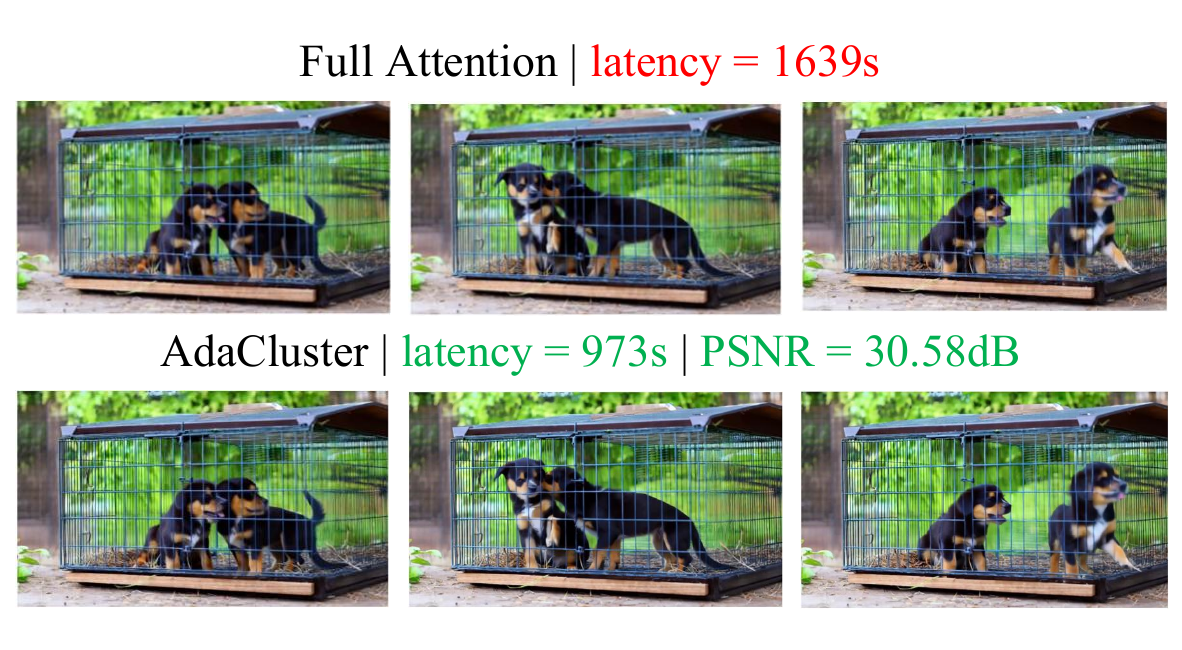}
    \caption{Performance comparison between Full Attention and \clustername methods on HunyuanVideo, with videos generated on a single A40. \clustername reduces inference latency from 1639s to 973s with a PSNR of 30.58dB, significantly improving inference speed while maintaining visual quality.}
    \label{fig:placeholder}
\end{figure}
The primary source of this cost is the attention module, whose complexity scales quadratically with the input sequence length~\cite{vaswani2017attention}. 
In video generation, the sequence length grows with both resolution and the number of frames, resulting in substantially longer input sequences.
For instance, in CogVideoX-2B~\cite{yang2024cogvideox}, we generated a video with 81 frames at a resolution of 720p on a single NVIDIA A40 GPU. The entire generation takes 1691 seconds, where the input sequence length is 70K tokens and attention accounted for 75\% of the total generation time.
We also observe similar phenomena in HunyuanVideo~\cite{kong2024hunyuanvideo} and other DiT models.

To address this, existing works, such as  Top-K attention~\cite{shao2022dynamictokennormalizationimproves,Zhang_2025_ICCV},  exploit the inherent sparsity of attention to accelerate computation, namely only a few critical tokens dominate output accuracy~\cite{zhu2019empirical,wang2021spatten}. One category of methods aggregates consecutive tokens into fixed-size blocks, and select a representative (usually the average of the tokens) from each block for significance estimation~\cite{zhang2025fast,zhang2025spargeattn}. However, consecutive tokens are not necessarily semantically close in the embedding space. Some tokens may be far away from the block representative, and hence leads to degraded attention accuracy.
% Since consecutive tokens may be semantically heterogeneous in the embedding space, such block-wise strategies can overlook critical tokens, leading to degraded attention accuracy. 
To address this issue, the token clustering methods are adopted in the later methods~\cite{yang2025sparse}, through which higher intra-group token similarity is ensured, hence achieving better accuracy.

% . Specifically, queries and keys are clustered into a fixed number of groups, and only the most salient clusters are selected for attention computation, thereby enhancing both flexibility and accuracy.

%In the clustering based methods, to achieve a significant speedup, both the queries and keys are clustered. However, the existing methods treat the query and key equally, and apply common clustering method to them. We analyzed that in the attention mechanism, the query and key play different roles, and they have markedly different patterns. Better performance can be achieved if customizing suitable clustering plan for them respectively. 

%In this paper, we propose \textbf{\sysname}, a training‑free adaptive clustering framework that designs customized clustering methods for query and key tokens respectively, yielding high generation efficiency and video quality. The core technical contributions are as follows.

In cluster-based sparse attention methods, queries and keys are typically clustered to achieve significant acceleration. However, existing approaches apply a uniform strategy to both queries and keys, using the same Euclidean distance-based clustering. This overlooks the distinct roles that queries and keys play in the attention mechanism, as well as the significantly different patterns they exhibit.

In this paper, we propose \textbf{\sysname}, a training-free adaptive clustering framework that introduces a role-aware redesign of the entire “cluster-then-select” pipeline. We design customized clustering strategies for queries and keys separately. The core technical contributions are as follows:
% : (1) we analyze that the relative magnitude of query–key scores is independent of query length. Based on this, \sysname normalizes queries and applies angle‑based clustering for higher compression. In contrast, the more heterogeneous distribution of key vectors requires euclidean‑level clustering, for which we design a multi‑stage adaptive strategy to preserve relationship between critical tokens and maintain attention accuracy. (2) To efficiently identify salient tokens after clustering, we introduce TensorQuest, a GPU Tensor‑Core friendly scoring mechanism that rapidly selects critical clusters. This design ensures that no important tokens are overlooked while substantially reducing computational overhead.

\begin{enumerate}
\item \textbf{Query Clustering.}
For the query tokens, we analyze that the relative magnitude of query–key scores is independent of the query length. Based on this, \sysname first normalizes queries and proposes an angle‑based clustering method, which can compress the query tokens by a large ratio. 
\item \textbf{Key Clustering.}
For the key tokens, we show that the distribution of the key tokens varies greatly across different layers. Due to this, we propose an layerwise adaptive kmeans clustering method, covering cluster number assignment, threshold-wise adaptive clustering, and efficient critical cluster selection.
\item \textbf{Evaluation.}
We build custom operators for \sysname on Triton~\cite{10.1145/3315508.3329973} and FlashInfer~\cite{ye2025flashinfer}, and test it on open-source DiT models including CogVideoX-2B~\cite{yang2024cogvideox}, HunyuanVideo~\cite{kong2024hunyuanvideo}, and Wan-2.1~\cite{wang2025wan}. Experimental results on one A40 GPU demonstrate that \sysname achieves an end-to-end acceleration of \textbf{1.67$\times$}–\textbf{4.31$\times$} for resolutions above 720p within our test range, 
while maintaining high visual fidelity with PSNR of 30.99. 
\end{enumerate}

% We build custom operators for \sysname on Triton~\cite{10.1145/3315508.3329973} and FlashInfer~\cite{ye2025flashinfer}, and test it on open-source DiT models including CogVideoX-2B~\cite{yang2024cogvideox}, HunyuanVideo~\cite{kong2024hunyuanvideo}, and Wan-2.1~\cite{wang2025wan}. Experimental results on one A40 GPU demonstrate that \sysname achieves an end-to-end acceleration of \textbf{1.67$\times$}–\textbf{4.31$\times$} for resolutions above 720p within our test range, 
% while maintaining high visual fidelity with PSNR of 30.99. 

%-------------------------------------------------------------------------

\section{Background and Motivation}
\label{sec:background}

\subsection{DiT Models and Computation Bottleneck}
\para{DiT models.} Video generation has emerged as a transformative technology, enabling the creation of high-quality, realistic videos for applications in entertainment, advertising, and virtual reality~\cite{wang2025surveyvideodiffusionmodels}. Recently, Video Diffusion Transformers (DiTs) have demonstrated superior performance in generating temporally consistent and visually compelling videos by leveraging a Transformer-based architecture~\cite{xing2024survey,wang2024av,li2024arlon,chen2024gentron,zhang2025tora}.

Popular DiT models share a similar architecture and inference workflow. Each layer typically includes an attention module to capture spatial-temporal dependencies across frames and a feed-forward network (FFN) for nonlinear feature enhancement~\cite{ma2024latte,xing2024survey}. During inference, DiTs generate videos by iteratively denoising random latent noise over dozens of steps, with each step requiring a full forward pass to ensure temporal coherence.

\para{Computation bottleneck of DiT.}  Despite their capabilities, generating high-quality videos with DiTs remains prohibitively time-consuming and resource-intensive. For instance, generating an 81-frame video at 1280$\times$720 resolution takes more than 27 minutes on an A40 GPU with 48GB HBM. A higher-resolution version (1920$\times$1120) of the same video takes drastically longer (130 minutes).

Our analysis identifies the attention mechanism as the primary performance bottleneck, a finding consistent with prior research~\cite{tsotsos2021computational,nagrani2021attention,zhao2024real}. In the two tasks mentioned above, attention computation accounted for 67.1\% and 83.8\% of the total time, respectively. This bottleneck arises because the computational complexity of attention scales quadratically with the input sequence length~\cite{shen2021efficient}. In DiTs, this sequence length is the product of the frame count, height, and width, leading to extremely long sequences~\cite{hacohen2024ltx}. For our examples, the sequence lengths reached 70K and 180K tokens, respectively. We observed a similar phenomenon in CogVideoX-2B~\cite{yang2024cogvideox}, another widely used DiT model. In summary, the performance bottleneck caused by excessively long sequences in DiT attention computation severely hinders the development of applications requiring high resolution or long video generation.

\subsection{Sparse Attention}

% \para{Sparse Attention.} 

To address the bottleneck of full attention, researchers have exploited the inherent sparsity of attention mechanisms~\cite{yoon2024exploring,yang2024post,deng2024sparse}. Only a small fraction of key/value tokens contribute significantly to the final output, so selectively retaining these important tokens can substantially reduce computational overhead while preserving model quality~\cite{an2024does,hua2022transformer}. Existing approaches generally fall into two categories: static sparsity and dynamic sparsity.

\textbf{Static sparsity} simplifies the sparsity identification during inference by using predefined and fixed attention patterns. 
% Methods such as MInference\cite{jiang2024minference} and SVG\cite{xi2025sparse} analyze the model’s attention behavior offline and assign fixed sparse patterns to each attention head based on empirical statistics or heuristic rules. These patterns typically emphasize spatial or temporal dependencies, reflecting the characteristics of attention heads in video diffusion transformers\cite{liu2021transformer,yang2025decision}. 
Methods such as MInference~\cite{jiang2024minference} and SVG~\cite{xi2025sparse} perform offline analysis of attention behavior and assign fixed sparse patterns to each attention head based on empirical statistics or heuristic rules. 
These sparsity patterns typically capture spatial or temporal dependencies, reflecting the inherent characteristics of attention heads in DiT~\cite{liu2021transformer, yang2025decision}.

However, static patterns lack sufficient generality and fail to capture all important attention regions, potentially omitting critical tokens. 
% As a result, mainstream approaches increasingly rely on Dynamic Sparsity Models\cite{wu2024sc4d,silveria2025chipmunk,zhang2025vsa,li2025magicmotion,xia2025training,tan2025dsv}.
To address this, recent methods increasingly resort to \textbf{dynamic sparsity} schemes~\cite{wu2024sc4d,silveria2025chipmunk,zhang2025vsa,li2025magicmotion,xia2025training,tan2025dsv,zhang2025spargeattn,yang2025sparse}.
% The Top-K attention method first computes the query-key scores $s=qK^T$ for each query, and then selects critical KV tokens with Top-K query-key scores for the final attention calculation. 
The most representative family is the Top-K attention approach, which computes the attention score matrix $S=QK^T$ and selects the Top-K key–value pairs via these scores, hence preserving enhanced flexibility and model quality.
However, computing the whole query-key scores for all the queries is very time-consuming.
% These methods construct attention masks by adaptively evaluating token importance from input features, enabling input-driven efficient computation with enhanced flexibility while preserving model quality.
% Dynamic sparsity methods, such as SpargeAttn\cite{zhang2025spargeattn} and SVG2\cite{yang2025sparse}, construct sparse attention masks by dynamically evaluating the importance of each token based on input features, achieving higher flexibility and computational acceleration while maintaining model performance. 
% Unlike static sparsity methods, dynamic sparsity does not rely on predefined fixed patterns; instead, it adaptively selects the attention regions to compute during inference, enabling input-driven efficient computation. 
To address this, SpargeAttn~\cite{zhang2025spargeattn} detects sparsity patterns at the block level by grouping consecutive tokens. 
In contrast, Sparse VideoGen2 (SVG2)~\cite{yang2025sparse} introduces token clustering for compression, yielding more flexible and efficient sparse attention scheduling.
Sparse attention variations with dynamic input patterns are more widely adopted due to their generality and flexibility, which is the focus of this paper.

\subsection{Limitations of Dynamic Sparsity Methods}
% Dynamic sparsity methods reduce computation by selectively computing attention for important tokens, but existing approaches face three critical limitations.
% Although existing methods dynamically select important tokens during inference, we observe that they still face three critical limitations.
Here, we conduct a study to uncover the limitations of existing dynamic sparsity and motivate our new design. To do so, we generate videos with Hunyuan~\cite{kong2024hunyuanvideo} and Wan-2.1~\cite{wang2025wan} models, and analyze the distribution of query-key vectors used for computing the attention scores. The main findings are summarized as follows.

\begin{figure}
    \hspace{0pt}
    \centering
    \includegraphics[width=1\linewidth]{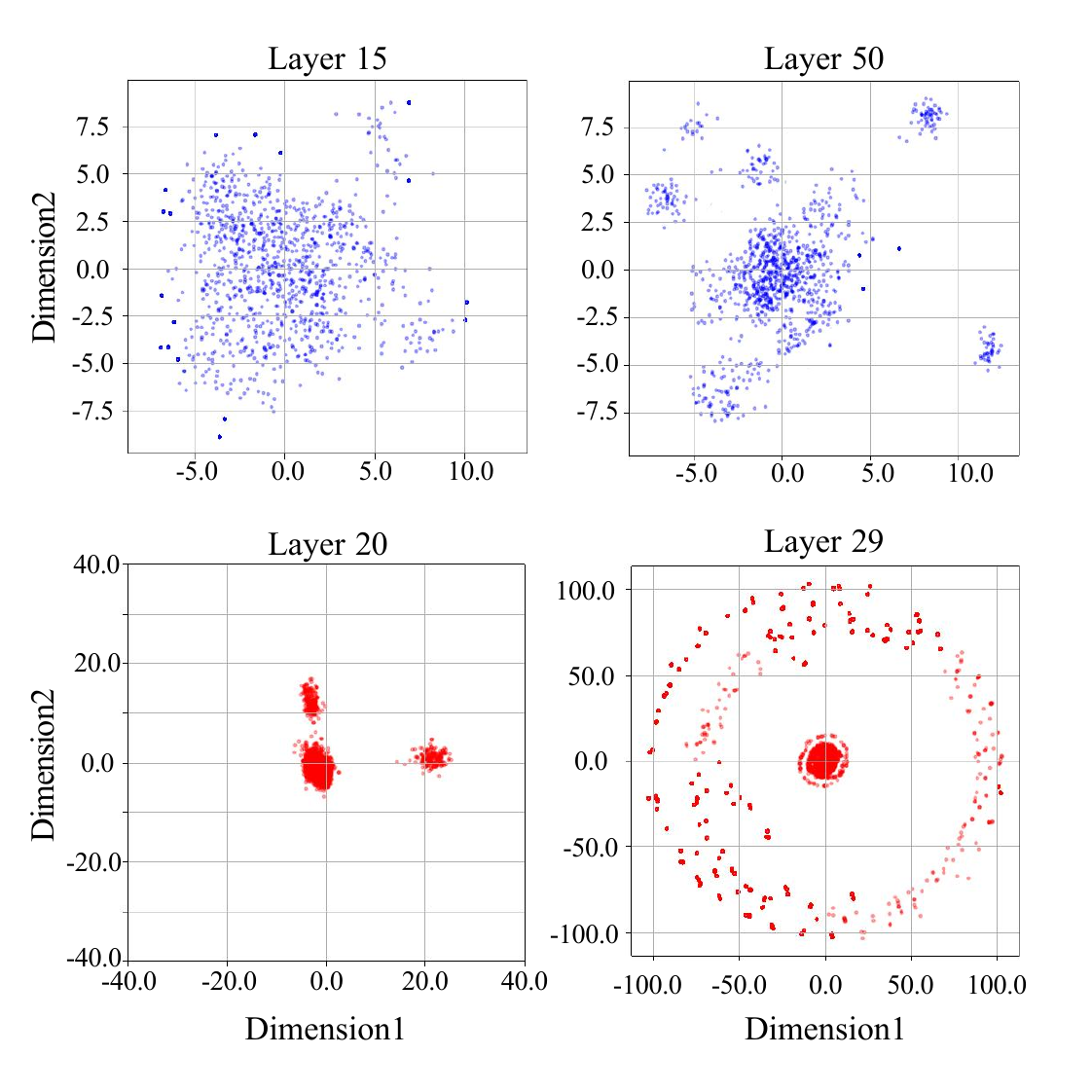}
    \caption{Token distribution visualization across different layers in Hunyuan(blue) and Wan-2.1(red) models. The token distributions are obtained by randomly sampling tokens across all steps of generating, and their 128-dimensional features are projected into a 2-dimensional space using PCA for visualization.}
    \label{fig:tokens}
\end{figure}

\para{Distributions of query-key vectors are ignored.} 
% Empirical evidence from our detailed analysis shows that different layers exhibit significant variations in the compactness of the token distribution. % (see Figure~\ref{fig:tokens}). 
SpargeAttn~\cite{zhang2025spargeattn} processes vectors in consecutive blocks without considering inherent properties such as numerical similarity. SVG2~\cite{yang2025sparse} considers similarity but uses a fixed clustering scheme with 100 query clusters and 500 key clusters for all models. 
However, we observe that the distribution of query–key vectors differs substantially across models and layers. Here, we take the numerical distribution of key vectors as an example.
As shown in Figure~\ref{fig:tokens}, some layers have highly dispersed token representations, requiring more clusters to preserve critical keys, or even being unsuitable for clustering. 
In contrast, layers with concentrated token distributions can be effectively compressed with fewer clusters.
% Moreover, query and key vectors exhibit distinct feature characteristics (see Sec~\ref{sec:systemdesign}), 
This diversity makes it difficult for existing static clustering methods to effectively capture critical tokens, potentially causing reduction in attention accuracy and output quality.
% Existing methods that apply a uniform clustering strategy across all layers fail to account for these differences, potentially causing over-compression in dispersed layers and under-compression in concentrated layers, which reduces attention accuracy and output quality.

\para{Critical tokens may be missed after clustering.} 
% Existing cluster-based selection methods typically rely on the attention weight of cluster centers to identify important clusters, which may result in some critical tokens being overlooked, thereby reducing the accuracy of subsequent attention computations.
After clustering, we need to evaluate the significance of all clusters for each query vector and select tokens from the Top-K critical clusters for attention computation. Existing methods typically estimate cluster significance based on attention scores computed over cluster centers. However, this approach may overlook important tokens, since a cluster center cannot fully represent all critical tokens, especially when those tokens lie near the cluster boundaries rather than close to the center.

% Together, these limitations highlight the requirement for adaptive query-key clustering methods that are layer-aware, aware of token-to-center distances, and temporally consistent to achieve efficient and high-fidelity video generation.
Together, these limitations underscore the need for an adaptive query–key clustering mechanism that can dynamically determine the appropriate number of clusters, and efficiently identify critical tokens after clustering. 
% Ultimately, the method should maintain temporal consistency to achieve efficient and high-fidelity video.

% ==============================
\section{Methodology}
\label{sec:systemdesign}

\begin{comment}
This section introduces the overall design of our token-efficient attention framework. The objective is to reduce the computational and memory cost of attention while maintaining semantic integrity and generation quality. To this end, we propose a layer-wise adaptive token clustering (AdaCluster) method which is able to efficiently select critical tokens for attention computing. AdaCluster assigns suitable number for clusters for different layers according to the specific token distribution, ensuring that the tokens within the same cluster are compact enough so that all the tokens can be approximated by their corresponding clusters. In addition, AdaCluster can further identify the hard-to-compress layers which requires vanilla full attention to preserve model accuracy. Since the number of such layers is relatively small, it will not significantly affect efficiency.
\end{comment}

This section introduces the overall design of AdaCluster. In the existing clustering based method SVG2~\cite{yang2025sparse}, both query and key vectors are treated equally, and they were clustered using the same manner. To maximize efficiency and meanwhile ensuring accuracy, AdaCluster designs customized clustering methods for the queries and keys, respectively.
We analyze that in the attention mechanism, the query and key have different patterns. 
In query clustering, we only need to ensure a relaxed angle similarity, so we normalize them before clustering, which can increase the compression ratio. Unlike this, the key vectors clustering is relatively complex, since it requires a critical euclidean similarity. To this end, we propose an effective method, covering cluster number assignment, threshold-wise adaptive clustering, and efficient critical cluster selection. 
% we first analyze the token distribution and compressibility across transformer layers to identify redundancy patterns\sn{To this end, for each layer, we first analyze the token distribution, identify the redundancy patterns, and propose a method to measure its compressibility}. Based on these observations, \sn{we select the layers with high compressibility score and compress these layers based on} a threshold-aware hierarchical K-means clustering algorithm that produces stable and consistent token groupings. \sn{Then we further propose a method to measure the importance of all the clusters and apply efficient Top-k attention to these layers by only using the tokens belonging to the critical clusters. For some hard-to-compress layers, we still apply the original full attention to preserve accuracy. Since the number of such layers is relatively small, it will not significantly affect efficiency.} 

% followed by an upper-bound approximation method for Top-K token selection that ensures all critical tokens are preserved with minimal overhead.

\subsection{Query Clustering}

\begin{figure}[t]
  \centering
  \includegraphics[width=1\linewidth]{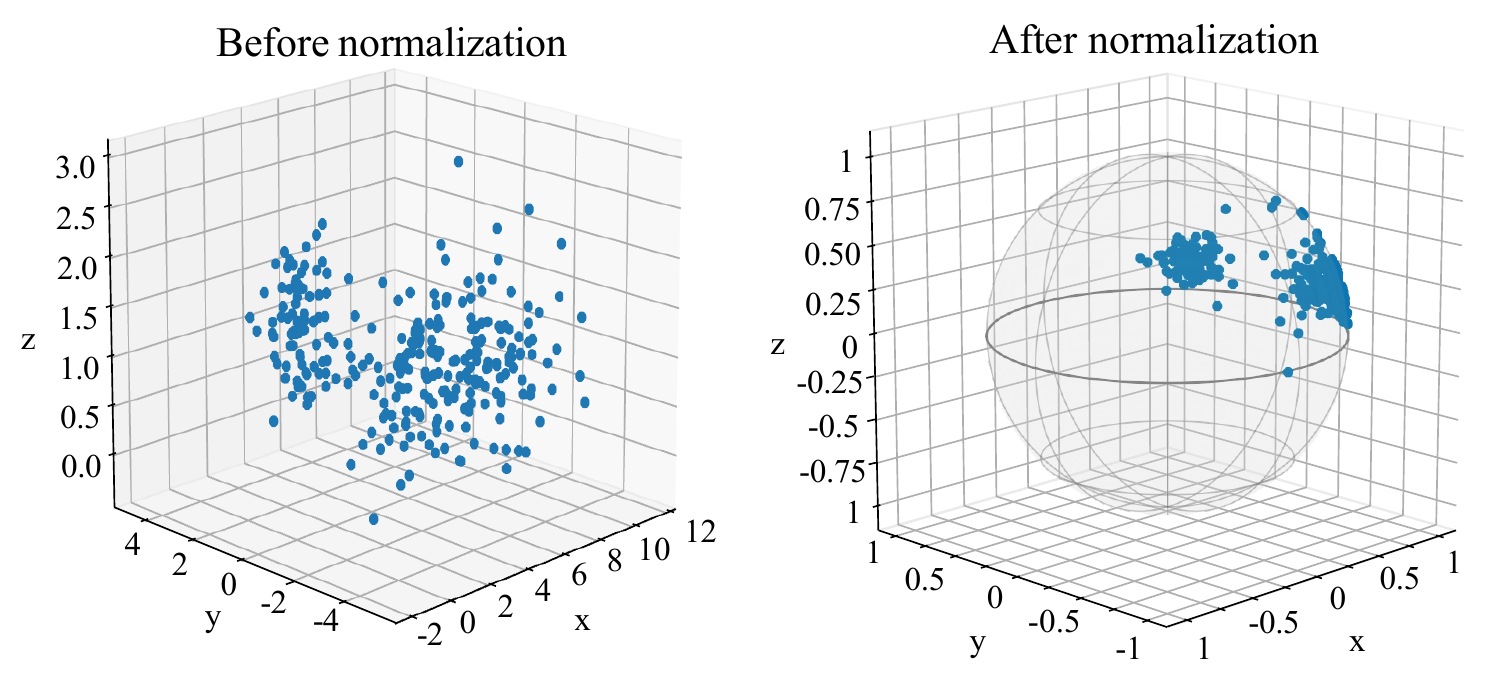}
  \caption{Query vector distributions before (left) and after (right) normalization. We normalize queries onto a unit sphere, making their distribution more compact and clustering more efficient.}
  \label{fig:q}
\end{figure}

% As stated above, we aim to compress both query and key vector by clustering, hence acceleration the attention computing in DiT denoising. 
We first introduce the method to compress the query vector. 
% Though SVG2~\cite{yang2025sparse} proposed to first cluster the queries, and use a small number cluster centers to select the critical KV tokens,  
Since the query vectors are distributed in the high-dimensional space, and the lengths of the vectors vary greatly, directly clustering on the original space is difficult, only limited compression ratio can be achieved. Note that for any query $q$, the relative magnitude of the  query-key scores $s$ is independent of the query vector length~\cite{2025Scale}. Hence, we can first normalize the query vectors, and use the normalized query to measure the significance of the key vectors. As shown in Figure~\ref{fig:q} , the distribution of the queries is much more compact, so we can cluster the queries by a much high compression ratio. Overall, in our query clustering method, we actually group the queries that are close at the angle level, which is much more efficient than the existing method using the Euclidean-level clustering.The corresponding theoretical analysis can be found in Appendix A.2. 
\subsection{Key Clustering}
\subsubsection{Observation and Analysis}  
\label{subsec:compressibility}
\begin{figure}[t]
  \centering
  \includegraphics[width=1\linewidth]{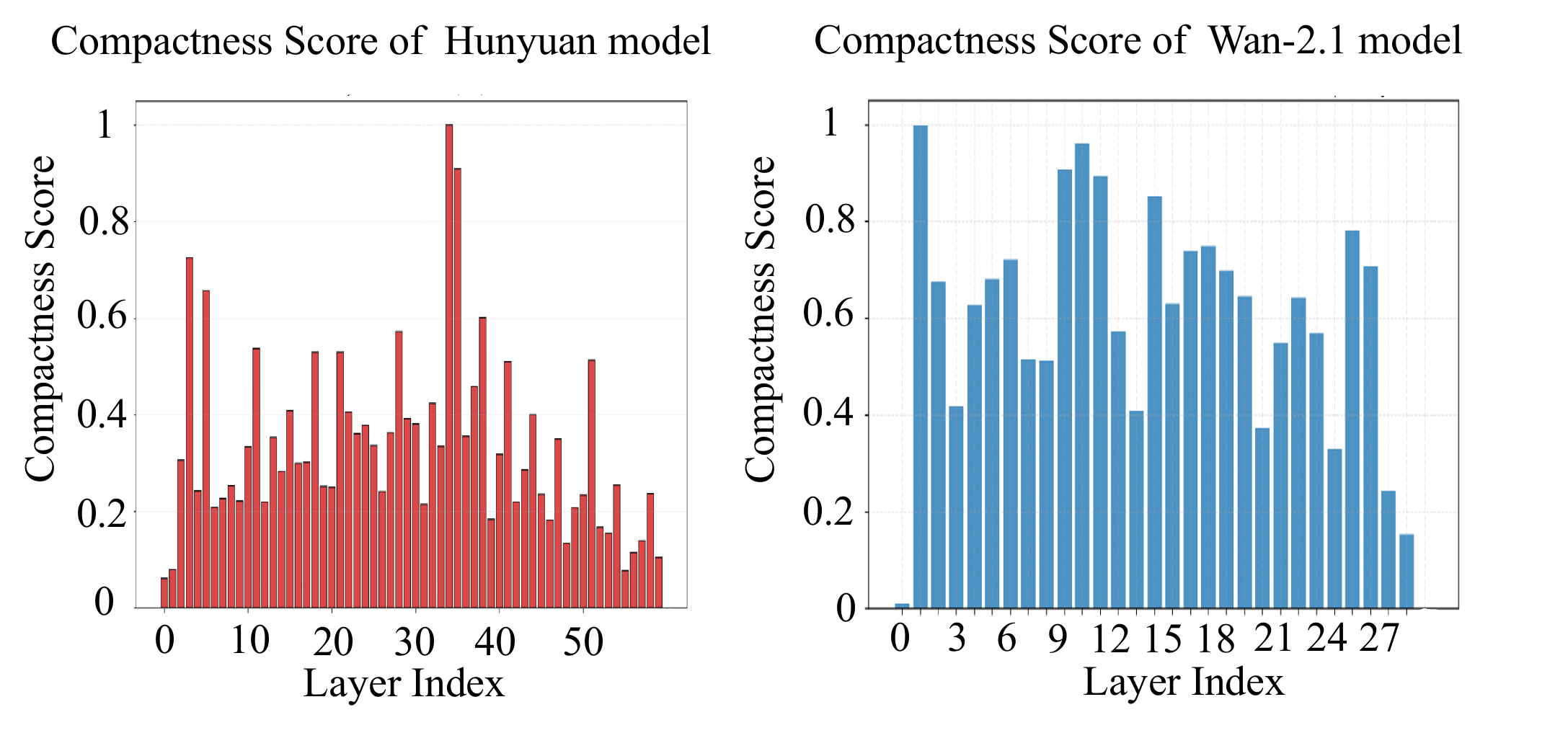}
  \caption{Layer-wise compactness scores of the Wan-2.1 and Hunyuan models. The x-axis represents the layer index, and the y-axis shows the normalized compactness score $C_l$. Note that the values are computed with respect to 400 cluster centers there.}
  \label{fig:compressibility}
\end{figure}
% As stated before, we aim to compress the attention layers using token clustering. The core principle of our clustering approach is to group similar tokens together, which enables efficient compression while preserving important information. The rationale is that if a cluster is close to the query $Q$, then all tokens within that cluster are likely to be important. In this context, a small mean squared error (MSE) becomes crucial, as it reflects the proximity of tokens within each cluster, ensuring that tokens assigned to the same cluster are indeed similar and can be effectively represented by their cluster center. The clustering distance plays an important role in preserving accuracy when finding the top-$K$ clusters, as it ensures that only clusters with sufficient proximity to the query are selected.
% \sn{Following svg \cite{}, we first cluster the tokens, then for each query we measure the significance of each cluster. We assume that the tokens in the same cluster are compact such that if a cluster is  critical according to the significance score for a target query, all the tokens belonging to this cluster will be used  for attention calculating. On the other hand, the tokens belonging to the clusters with low  significance scores will be directly discarded.  }
Though the complexity of attention is greatly reduced after query clustering, when the sequence length is long, identifying significant key tokens is still the most time-consuming step in Top-K attention.  To further speed up the denoising process, we also consider to cluster the key vectors. Different from the queries, both the vector direction and the vector lengths of the keys affect the Top-K attention result, so the keys require Euclidean-level clustering.  Considering an arbitrary $k$, assume $c(k)$ is the cluster center of $k$ after token clustering. Note that for any $q$, we have that
\begin{equation*}
    q^\top c(k) - q^\top k \leq ||q|| * ||k-c(k)||,
\end{equation*}
so if the tokens in the same cluster are compact enough, namely $||k-c(k)||$ is very small, then we can efficiently approximate the qk score $q^\top k$ by $q^\top c(k)$, for all the keys in this cluster. In this situation, for any query $q$, we do not need to measure and compare the significance of all the keys. Instead, we only need to measure and compare the significance of clusters.

According to above analysis, the compactness of the key vectors in the same cluster determines whether the critical keys can be accurately select, and hence affects the final model performance. SVG2 adopts uniform number of clusters for all the layers. However, we observe that the range of keys distribution varies greatly across different layers, as show in Figure~\ref{fig:tokens}. To measure the intra-cluster compactnessof different layers more explicitly, we randomly sample a request with length $N$, and cluster the key tokens of all the layers with a uniform number of cluster centers. 
Then we compute the mean squared reconstruction error (MSE) of each head $i$ in layer $l$ by
\begin{equation*}
    \mathrm{MSE}^i_l = \frac{1}{N} \sum_{i=1}^{N} \big\| k^i_l - c({k}^i_l) \big\|_2^2
\end{equation*}
where $k^i_l$ denotes the $i$-th key vector of head $i$ from layer $l$. We define the compactness score of each layer $l$ as $\mathrm{Comp_l} = 1/\mathrm{MSE_l}$, where $\mathrm{MSE}_l$ is the average of  $\{ \mathrm{MSE}^i_l\}$. 

Figure ~\ref{fig:compressibility} measures the compactness scores of different layers for model  Wan-2.1 and Hunyuan. It is shown that, the intra-cluster compactness differs significantly across different layers.  Hence, we should assign larger cluster counts for the layers with relatively scattered data distribution, and less cluster counts for the layers with relatively concentrated data distribution. The compressity for different prompts within the same layer is similar (Appendix A.1). Therefore, we only need to adjust on a per-layer basis.

\subsubsection{Multi-stage K-means Clustering}
\begin{figure*}[t]
    \centering
    \includegraphics[width=0.85\linewidth]{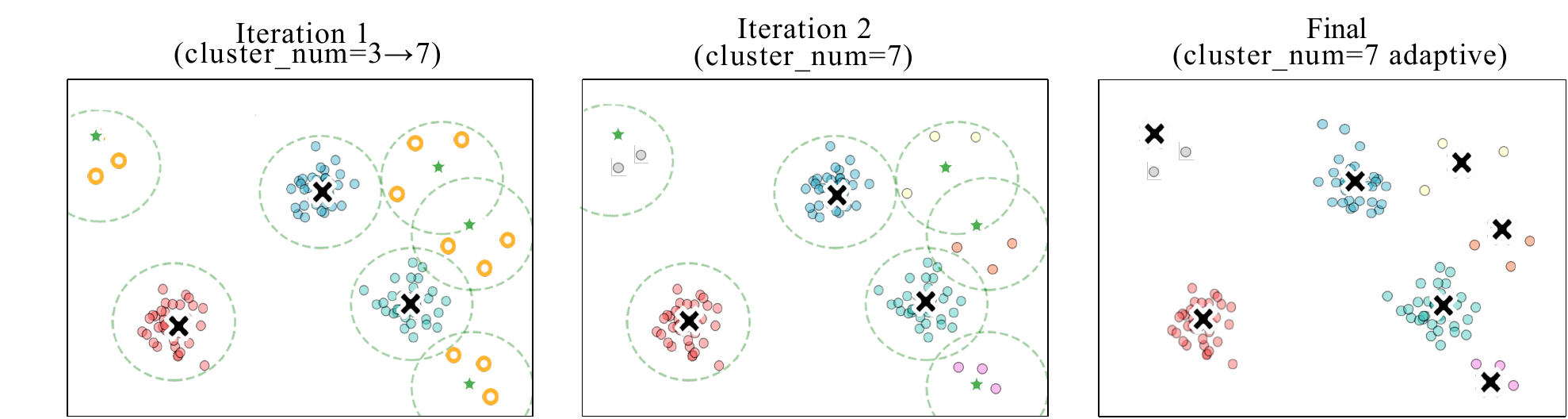}
    \caption{Visualization of Multi-stage K-means clustering. Red crosses represent tokens, black crosses denote cluster centers, and dashed circles indicate cluster boundaries. In each iteration, samples are progressively clustered into finer groups. The number of clusters (\textit{cluster\_num}) increases adaptively as unassigned samples are reclustered. }
    \label{fig:iterative_clustering}

\end{figure*}

As analyzed above, different number of clusters are required across layers to ensure the compactness of intra-cluster data distribution. However, it is difficult to determine the suitable number of clusters in advance. In this subsection, we adopt a multi-stage K-means clustering method to finish this task. Specifically, as shown in Figure ~\ref{fig:iterative_clustering}, for each layer, at the initial stage, we cluster the tokens with a moderate number of clusters. After clustering, we select the outlier tokens which is relatively far from their cluster centers (beyond a predefined threshold). Then we re-cluster these outlier tokens with a few number of new clusters. We repeat such a strategy until each token is assigned to a compact cluster. In particular, if the number of clusters exceeds a preset upper limit, we will stop further clustering and treat this layer as a hard-to-compress layer. For such layers, we apply the vanilla full attention. The detailed process is shown in Algorithm ~\ref{alg:threshold_clustering}.

\begin{algorithm}
\caption{MultiStageClustering}
\label{alg:threshold_clustering}
\begin{algorithmic}[1]
\Require Key vectors set $\mathcal{K} = \{\mathbf{k}_1, \dots, \mathbf{k}_n\}$, distance threshold $\tau$, max cluster count $N_{\text{max}}$
\Ensure Cluster assignments $\mathcal{A}$, Cluster centers $\mathcal{C}$, use full attention flag $F$
\State $\mathcal{C} \gets \emptyset$
\State $\mathcal{U} \gets \mathcal{D}$ 
\State $t \gets 0$, $F \gets \text{False}$

\While{$\mathcal{U} \neq \emptyset$}

    \If{$|\mathcal{C}| \ge N_{\text{max}}$}
        \State $F \gets \text{True}$ 
        \State \textbf{break}
    \EndIf

    \State Choose a moderate cluster number $m_t$
    
    \State $\text{Index}, \mathcal{C}_t \gets \text{KMeans}(\mathcal{U}, m_t)$
    
    \For{each $\mathbf{k}_i \in \mathcal{K}$}
        \State Compute the distance $d_i \gets \|{k}_i - c(k_i)\|_2$
        \If{$d_i < \tau$} 
            % \State $\mathcal{A}(\mathbf{x}_i) \gets \arg\min_{c \in \mathcal{C}_t} \|\mathbf{x}_i - \mathbf{c}\|_2$
            \State Remove $k_i$ from $\mathcal{U}$: $\mathcal{U} \gets \mathcal{U} \setminus \{k_i\}$
        \EndIf
    \EndFor
    
    \State $\mathcal{C} \gets \mathcal{C} \cup \mathcal{C}_t$

    \State $t \gets t + 1$
\EndWhile

\State Obtaining the cluster assignment set $\mathcal{A}$ by assigning each $k_i$ in $\mathcal{K}$ to the closest cluster center in $\mathcal{C}$.
\State \Return $\mathcal{A}, \mathcal{C}, F$
\end{algorithmic}
\end{algorithm}

\begin{figure}[t]
  \centering
  \includegraphics[width=1\linewidth]{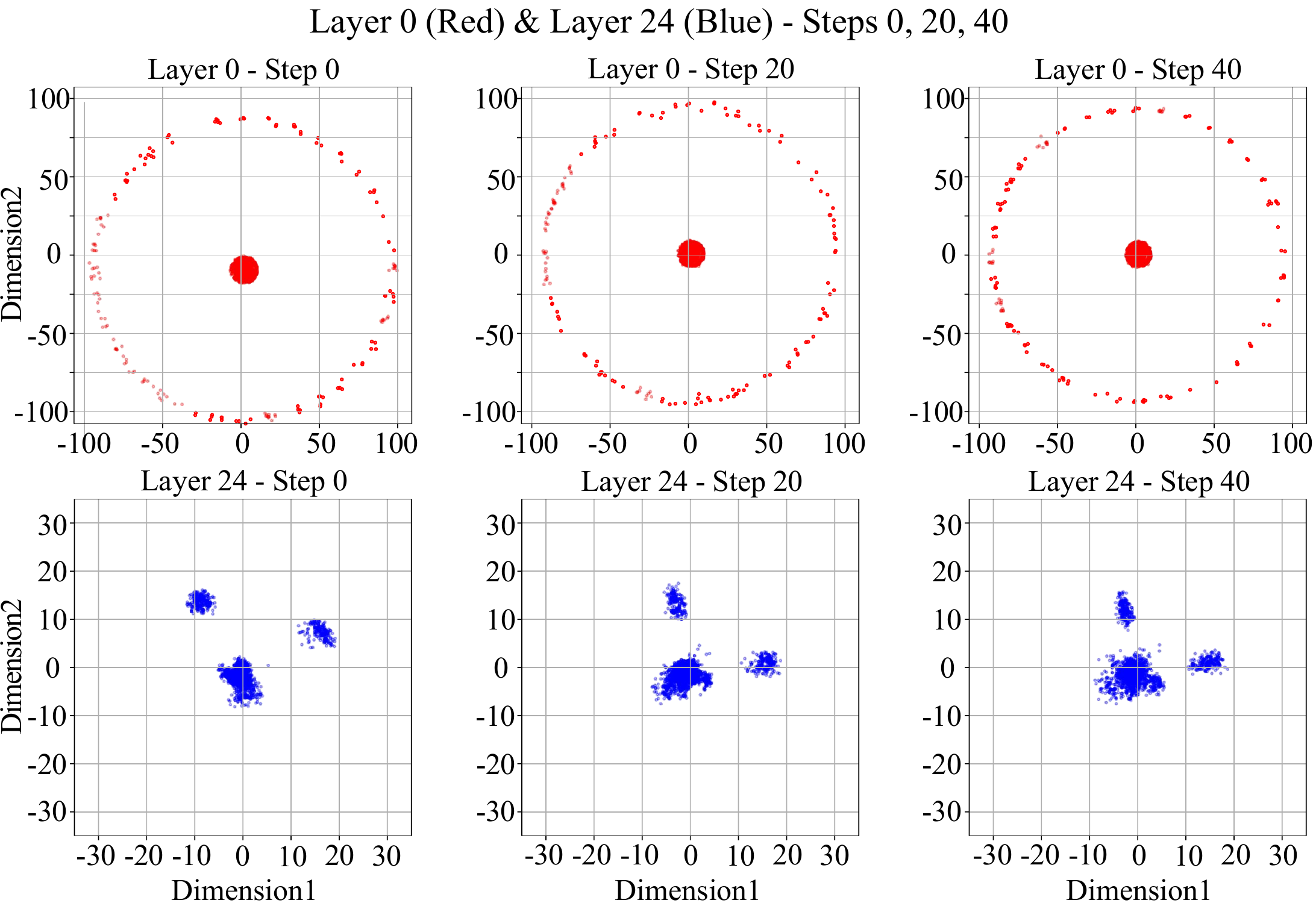}
  \caption{Token distribution consistency of Wan-2.1 across timesteps for Layer 0 (top, red) and Layer 24 (bottom, blue). Each column corresponds to a different denoising step (0, 20, 40). }
  \label{fig:layer_step_grid}
\end{figure}

\para{Efficient initialization.}
DiT involves multiple denoising steps during image/video generation. We observe that for each layer the token distributions change gradually across consecutive denoising steps, i.e., adjacent steps exhibit very similar distributions. Figure~\ref{fig:layer_step_grid} visualizes token distributions across denoising steps (Layer 0 and Layer 24) using PCA. Therefore, the cluster centers computed at step $t$ can approximately represent the token distribution of step $t{+}1$.

% We further observed that token distributions within the same layer change gradually across consecutive timesteps, i.e., adjacent steps exhibit very similar distributions. 
% As a result, K-means cluster centers computed at step $t$ remain highly representative for step $t{+}1$, which naturally enables cross-step cluster center reuse. 

% Figure~\ref{fig:layer_step_grid} visualizes token distributions across denoising steps (Layer 0 and Layer 15) using PCA. Tokens maintain stable layouts and clustering patterns across steps, confirming the validity of reusing cluster centers for adjacent timesteps.

Based on this observation, we can further simplify the whole process of DiT denoising. We only need to apply the above-mentioned multi-stage kmeans clustering at the first denoising step to determine the number of clusters for each layer. In the later denoising steps, the  number of clusters is fixed, the specific cluster centers will be updated. In particular, we will use the cluster centers of previous denoising step as the initialization of the current step, which can accelerate the clustering procedure  in practice. 
% we fix the number of clusters $K_l$ for each layer across all denoising steps and reuse centers $\{\mathbf{c}_1^{(t)}, \dots, \mathbf{c}_K^{(t)}\}$ from step $t$ as initialization for step $t{+}1$.  
% This temporal consistency reduces redundant computation and improves feature alignment across timesteps, enhancing temporal coherence in generated videos.

% ------------------------
% ------------------------
\subsubsection{Efficiently Identify Critical Clusters}
\label{sec:tensorquest}
% Next, we need to measure the significance of all the clusters for any given query vector, and then we can select the tokens from the critical clusters for attention calculation. Existing cluster-based compression methods typically use the attention weights across the cluster centers to measure the significance of the clusters. Such a manner may miss critical tokens, especially when the critical tokens are located at the edge of the cluster rather than close to the cluster center. 
To efficiently identify critical tokens after clustering, we take inspiration from the existing method Quest~\cite{QuestAttn}, which is able to select the most potentially critical KV cache pages for the current query during LLM decoding. 
Specifically, given a query $q$, and a page of key vectors $K:=[k_1, k_2,...,k_n]$, Quest identifies critical attention candidates by estimating the upper bound of attention weights across all key vectors, which is expressed as 
\begin{equation} \label{eq:1}
    \text{Quest}(q, K) = \sum_{d=1}^D \max(q^d *\max(K^d), q^d*\min(K^d)),
\end{equation}
where $D$ is the total dimension of $q$ and $k$, and $K^d$ denote the vector composed of the $d$-th dimension of K.
One can refer to ~\cite{QuestAttn} for more details about Quest. 

Inspired by Quest, we aim to adopt a similar method to measure the significance of a cluster. However, according to (\ref{eq:1}) we can see that the Quest score is purely computed in the GPU CUDA Core, which is inefficient. Compared with LLM decoding, the attention in diffusion denoising is more computationally intensive, so the existing Quest method cannot be directly applied, due to unacceptable time delay. 
In this section, we propose TensorQuest, which is equivalent to vanilla Quest method, but enables the main computation to be executed in the powerful tensor core, thereby significantly enhancing efficiency. To achieve this goal, we first extract the positive and negative parts of query and key separately, as follows,
\begin{eqnarray*}
    \mathbf{q}^{+} &=& \max(\mathbf{q}, 0), \quad \mathbf{q}^{-} = \min(\mathbf{q}, 0), \\
    \mathbf{k}^{+} &=& \max(\mathbf{k}, 0), \quad \mathbf{k}^{-} = \min(\mathbf{k}, 0).
\end{eqnarray*}
Then the Quest score can be computed by
\begin{equation}\label{eq:2}
     \text{Quest}(q, K) = \text{matmul}(q^+, k^+) + \text{matmul}(q^-, k^-).
\end{equation}
It can be verified that equation (\ref{eq:2}) is equivalent to (\ref{eq:1}). Except the light-weight positive and negative parts extraction, the main computation is finished in the CUDA Core. We illustrate the whole process in Algorithm~\ref{alg:topk_tensorcore}.
\begin{algorithm}[t]
\caption{TensorQuest}
\label{alg:topk_tensorcore}
\begin{algorithmic}[1]
\Require 
\Statex Query tensor $\mathbf{Q} \in \mathbb{R}^{B \times H \times L_q \times D}$
\Statex Key tensor $\mathbf{K} \in \mathbb{R}^{B \times H \times L_k \times D}$
\Ensure Quest score $\mathbf{S}$ 

\State $\mathbf{Q}^{+} \gets \max(\mathbf{Q}, 0)$ 
\State $\mathbf{Q}^{-} \gets \min(\mathbf{Q}, 0)$ 
\State $\mathbf{K}^{+} \gets \max(\mathbf{K}, 0)$ 
\State $\mathbf{K}^{-} \gets \min(\mathbf{K}, 0)$ 

\State $\mathbf{S} \gets \mathrm{matmul}(\mathbf{Q}^{+}, (\mathbf{K}^{+})^{\top}) + \mathrm{matmul}(\mathbf{Q}^{-}, (\mathbf{K}^{-})^{\top})$

\Statex \quad \textit{// shape: $[B, H, L_q, L_k]$}

% \State $\mathcal{I} \gets argtopk(\mathbf{S}, k, \text{dim}=-1)$

\State \Return $\mathbf{S}$ 
\end{algorithmic}
\end{algorithm}

\begin{algorithm}[H]
\caption{AdaCluster}
\label{alg:adacluster}
\begin{algorithmic}[1]
\Require Query $Q$, Key $K$, Value $V$, distance threshold $\tau$, Top-K parameter $topk$, max cluster count $N_{\max}$
\Ensure AttnOut

\State $A_q, C_q \gets \text{KMeans}(\text{Normalize}(Q))$
\State $\mathcal{A}_k, \mathcal{C}_k, F \gets \text{MultiStageClustering}(K, \tau, N_{\max})$

\If{$F$ is True}
    \State $\text{AttnOut} \gets \text{Attention}(Q, K, V)$
\Else
    \State $S \gets \text{TensorQuest}(Q, K)$
    \State Index $\gets \text{argtopk}(S,  topk)$
    \State $K^* \gets K[\text{Index}], V^* \gets V[\text{Index}]$
    \State $\text{AttnOut} \gets \text{Attention}(Q, K^*, V^*)$
\EndIf

\State \Return AttnOut

\end{algorithmic}
\end{algorithm}

% The upper-bound guarantee ensures that no important clusters are missed during selection, maintaining high recall while enabling safe pruning of low-scoring clusters. 
\subsection{Overall Process}
% AdaCluster performs adaptive query-key clustering for efficient sparse attention in transformer layers. For each layer, it first estimates the appropriate number of query clusters based on the normalized query distribution and designates a small proportion of layers with high cluster counts as full-attention layers. For the remaining clusterable layers, queries are normalized and clustered via K-means, while keys are clustered using a multi-stage, threshold-aware procedure to ensure compactness. TensorQuest computes upper-bound attention scores using positive and negative parts of queries and keys, selecting the Top-K key clusters for each query. Attention is then computed only over the selected key clusters, and outputs from all layers are concatenated. Across consecutive denoising steps, cluster centers are reused as initialization to exploit temporal stability, further accelerating computation while preserving accuracy.
Finally, we describe the whole procedure of AdaCluster in the Algorithm \ref{alg:adacluster}. We cluster normalized queries using standard KMeans, while the keys are processed with our MultiStageClustering to determine whether a given layer should be compressed. For layers identified as compressible, we employ TensorQuest to rapidly select critical tokens for attention computation.

\section{Experiment}

\subsection{Setup}
\begin{table*}[t]
    \centering
    \setlength{\tabcolsep}{4pt}  
    \caption{Similarity, quality and efficiency benchmarking results of \sysname and baselines.}
    \begin{tabularx}{\linewidth}{
        >{\hsize=1.5\hsize\centering\arraybackslash}X 
        >{\hsize=1.01\hsize\centering\arraybackslash}X |
        >{\hsize=0.64\hsize\centering\arraybackslash}X |
        >{\hsize=0.64\hsize\centering\arraybackslash}X |
        >{\hsize=0.65\hsize\centering\arraybackslash}X |
        >{\hsize=0.93\hsize\centering\arraybackslash}X |
        >{\hsize=0.93\hsize\centering\arraybackslash}X |
        >{\hsize=0.99\hsize\centering\arraybackslash}X |
        >{\hsize=0.95\hsize\centering\arraybackslash}X |
        >{\hsize=0.76\hsize\centering\arraybackslash}X
    }
    \hline
         Model&  Config&  PSNR↑&  SSIM↑&  LPIPS↓& ImgeQual↑& BgConsis↑& SubConsis↑& TmpFlick↑& Speedup↑\\\hline
         CogVideoX-2B&  720$\times$480&  -&  -&  -&  61.15\%& 96.06\%& 92.71\%&  97.31\%&  1$\times$\\\hline
         &  SpargeAttn&  28.189&  0.517&  0.618&  55.77\%& \textbf{95.56}\%& \textbf{94.35\%}&  \textbf{98.14\%}&  1.23$\times$\\
         \rowcolor{blue!10}& AdaCluster&  \textbf{30.989}&  \textbf{0.767}&  \textbf{0.231}&  \textbf{56.76\%}& 93.95\%& 89.26\%&  95.18\%&  \textbf{1.67$\times$}\\\hline
         Wan-2.1-1.3B& 832$\times$480& -& -& -&  67.53\%& 96.72\%& 95.08\%&  98.04\%& 1$\times$\\\hline
         & SpargeAttn& 28.292& 0.437& 0.599& 64.59\%& \textbf{97.24\%}& \textbf{96.29\%}& \textbf{98.36\%}&  1.81$\times$\\
         & SVG2& 28.230& 0.358& 0.679& 66.43\%& 96.05\%& 94.36\%& 97.96\%&  1.61$\times$\\
         \rowcolor{blue!10}& AdaCluster& \textbf{29.083}& \textbf{0.571}& \textbf{0.393}& \textbf{66.45\%}& 96.35\%& 94.58\%& 97.84\%&  \textbf{1.85$\times$}\\\hline
         HunyuanVideo& 1280$\times$720& -& -& -& 62.64\%& 95.84\%& 93.05\%& 98.64\%& 1$\times$\\\hline
         & SpargeAttn& 28.155& 0.490& 0.596& 61.76\%& 95.89\%& \textbf{92.92\%}& 98.69\%& 1.33$\times$\\
         & SVG2& 29.319& 0.794& 0.308& \textbf{65.42\%}& \textbf{96.58\%}& 92.14\%& 98.73\%& 1.57$\times$\\
         \rowcolor{blue!10}& AdaCluster& \textbf{30.580}& \textbf{0.835}& \textbf{0.203}& 65.11\%& 95.58\%& 92.56\%& \textbf{98.85\%}& \textbf{1.68$\times$}\\\hline
    \end{tabularx}
    \label{tab:quality-and-speedup}
\end{table*}

\textbf{Models.} We evaluate the performance of \sysname on several Diffusion Transformer (DiT) based video generation models, including HunyuanVideo~\cite{kong2024hunyuanvideo, hunyuanvideo_repo}, Wan-2.1-T2V-1.3B~\cite{wang2025wan, wan_repo}, and CogVideoX-2B~\cite{yang2024cogvideox, cogvideo_repo}. We generate videos with multiple resolutions (e.g. 480p  and 720p) via the Text-to-Video (T2V) method.

\noindent\textbf{Datasets and Metrics.} For the generation task, we use the prompt dataset from PenguinVideoBenchmark~\cite{penguin}. To evaluate the performance of the \sysname, we measure the generated videos from two dimensions: similarity to original videos and overall video quality, with specific metrics selected as follows. For the similarity assessment between generated videos and reference videos, we choose three metrics: Peak Signal-to-Noise Ratio(PSNR), Learned Perceptual Image Patch Similarity(LPIPS) and Structural Similarity Index Measure(SSIM). For the overall video quality evaluation, we use VBench~\cite{Huang_2024_CVPR} to measure four aspects of generated videos:imaging quality, background consistency, subject consistency and aesthetic quality.

\noindent\textbf{Baselines.} We compare the performance of AdaCluster with baseline models adopting FlashAttention~\cite{dao2023flashattention2}. Additionally, we select SpargeAttn~\cite{zhang2025spargeattn} and SVG2~\cite{yang2025sparse} as baselines, which represent the state-of-the-art sparse attention algorithms with dynamic input-driven patterns. We use the default hyperparameter settings reported in their original papers and code repositories.

\noindent \textbf{Platform and configurations.} Main experiments are conducted on one NVIDIA A40 GPU with 48GB memory. Experiments on H100 are in Apendix A.6. Our implementation builds upon Transformer Diffuser~\cite{feng2023diffuser} and integrates custom kernels from FlashInfer~\cite{ye2025flashinfer}. For skipped layers, we use FlashAttention~\cite{dao2023flashattention2} as the full attention, while for layers with clustering, we implement a Triton-based kernel to improve computational efficiency. To balance speed and accuracy, we set $k_{\text{max}}$ such that the top 15\% of layers use full attention—these layers are relatively small, making the additional latency acceptable. The threshold $\tau$ is defined as 1.5$\times$ of the average token-to-cluster-center distance computed in the first inference step. We fix the number of clusters for q to 65, while the number of clusters for k is dynamically adjusted according to our algorithm. We report the detailed number of clusters for k under different videos configuration in the Appendix. Hyperparameter sensitivity analysis is in Appendix A.5. Due to memory constraints of A40, we use 30 inference steps for Hunyuan, while setting 50 steps for the other two models. Based on this hardware setup, we ensure that all baselines use the exact same model. For all subsequent timesteps, the cluster centers from the previous step are reused.

%jc \noindent\textbf{Implementation.} The experimental equipment is an NVIDIA A40 GPU, and we select SVG2 and SpargeAttention as baselines. The hyperparameters of the experiments are as follows: The $T_{\text{max}}$ in Algorithm 1 is set to 10, yet the number of iterations rarely reaches 10 during actual inference; $\tau$ is set to 1.5 times the average distance from tokens to their cluster centers during the first inference step. 

%jc \noindent\textbf{Baselines.} SVG2 and SpargeAttn are two representative baselines focused on efficient video generation, both designed to optimize attention computation to balance speed and quality. We compare our method with SVG2, SpargeAttn, and videos generated from the official repositories of each model, with all baselines adopting their respective official configurations.

\subsection{Quality Evaluation} We present all the accuracy evaluation results in Table ~\ref{tab:quality-and-speedup} along two dimensions: similarity to the full-attention model and video quality scores measured by VBench.

In terms of similarity, our results indicate that \sysname consistently outperforms the selected baselines across all three models. This demonstrates that \sysname effectively leverages diverse sparsity patterns across different model layers, thereby preserving video quality to a greater extent relative to the original full-attention approach.

Regarding video quality evaluation, we observe that \sysname and the two baseline methods generally achieve performance comparable to that of the original model. However, two exceptions are noted: first, SpargeAttn occasionally achieves an unexpectedly high score—sometimes even surpassing the original model—which may be attributed to its tendency to generate videos that align more closely with VBench's evaluation standards, thereby making such scores less meaningful in comparative analysis. 

Second, accuracy is stable on Wan-2.1 and Hunyuan with little variation across methods. On CogVideoX, however, both SpargeAttn and \sysname show a significant drop in image quality; SpargeAttn’s sub-consistency is much higher than the original, while ours is much lower. We attribute this to the model itself—CogVideoX exhibits the poorest video quality among the three (see Appendix). 

It is worth noting that video quality assessment remains an active area of research. Our objective here is primarily to demonstrate that our approach preserves visual quality without degradation, while simultaneously delivering substantial generation speedup. As shown in Table~\ref{tab:quality-and-speedup}, \sysname achieves the highest end-to-end generation speedup while incurring the lowest accuracy loss. Screenshots of the generated video are provided in the Appendix.

% \jc{We also report the end-to-end generation speedup ratio in Table~\ref{tab:quality-and-speedup}. The results show that \sysname achieves the highest speedup while incurring the lowest accuracy loss.}

% We test the video accuracy under different resolutions and evaluate the videos using Vbench metrics (ImageQual, AestheticQual) and similarity measures (PSNR, SSIM, LPIPS). The results show that \sysname outperforms the selected baselines (SVG2 and SpargeAttention) in most cases, with the comparison results presented in Table ~\ref{tab:quality-and-speedup}.

% In terms of similarity metrics, our method outperforms the two selected baselines across the board, showing significant advantages in both SSIM (higher is better) and LPIPS (lower is better), while achieving superior PSNR (higher is better).

% For video accuracy, We observe that our method and the other baselines all exhibit quality comparable to full attention. However, we identified two exceptions: one is the unstable scores on Cog, which is attributed to inherent quality issues of the Cog model itself; the other is occasional instances of higher scores than full attention, which may stem from limitations of the benchmark itself. On the other hand, benchmarks for video quality are themselves an ongoing research topic. Here, we simply aim to emphasize that our method does not lead to quality loss, while simultaneously achieving significant acceleration performance.

\begin{figure}[t!]
    \centering
    \includegraphics[width=0.85\linewidth]{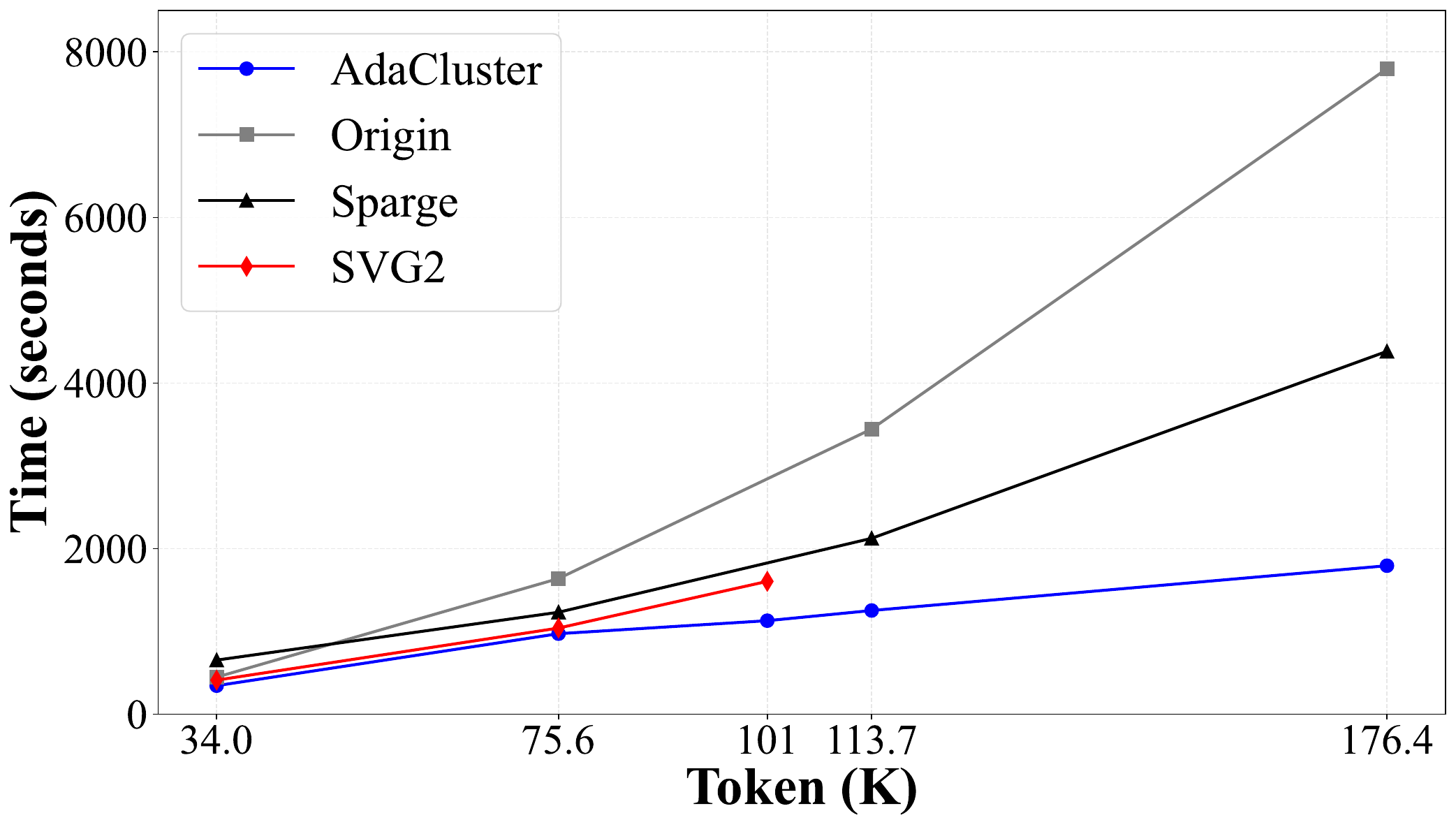}
    \caption{Time cost trend of several methods vs. seqlen.
    % \jc{Update the figure, remove 2.3K}
    }
    \label{fig:efficiency}
\end{figure}
\label{sec:evaluation}
\subsection{Efficiency Evaluation} In this section, we present the end-to-end generation time for various resolutions using HunyuanVideo as the test model. The number of generated frames is fixed at 81, while the resolution is varied, with the corresponding token counts reported in Table~\ref{tab:resolution}. As shown in Figure~\ref{fig:efficiency}, we observe that under different input token lengths, \sysname consistently achieves the highest speedup ratio. For instance, with 28.3K tokens, \sysname attains a speedup of 1.53$\times$, while SpargeAttn and SVG2 achieve 1.28$\times$ and 1.46$\times$, respectively. As the token count increases, the speedup of \sysname becomes even more pronounced. At the maximum input length of 176.4K tokens, \sysname and SpargeAttn reach their peak speedups of 4.31$\times$ and 1.78$\times$, respectively. This trend can be attributed to the inherent sparsity of the attention mechanism—longer sequences tend to exhibit greater sparsity, thereby allowing more room for acceleration. Additionally, we note that SVG2 is only applicable when the token count is below 101.1K, due to its substantial memory overhead from caching metadata for subsequent processing. 
%jc \youhuic{explain the speedup trends from short to long sequence.}

\begin{table}[t!]
    \centering
    \small
    \setlength{\tabcolsep}{2pt}
    \caption{Number of tokens corresponding to different resolutions for HunyuanVideo under the 81-frame configuration.}
    \begin{tabular}{@{}cccccc@{}}
    \toprule
    RES & 864$\times$480 & 1280$\times$720 & 1712$\times$720 & 1912$\times$720 & 1912$\times$1120 \\ \midrule
    Tokens & 34.0K    & 75.6K     & 101.1K    & 113.7K    & 176.4K     \\ \bottomrule
    \end{tabular}
    % \small
    % \setlength{\tabcolsep}{7pt}
    % \begin{tabularx}{\linewidth}{
    %     >{\centering\arraybackslash}X
    %     >{\hsize=0.9\hsize\centering\arraybackslash}X 
    %     >{\hsize=0.85\hsize\centering\arraybackslash}X 
    %     >{\centering\arraybackslash}X 
    %     >{\centering\arraybackslash}X 
    %     >{\centering\arraybackslash}X
    % }
    % \toprule
    %      Resolution&   864$\times$480&  1280$\times$720& 1712x720&  1920$\times$720& 1920$\times$1120\\\midrule
    %      Token(K)&  34.0&  75.6&  101.1&  113.7& 176.4\\\bottomrule
    % \end{tabularx}
    \label{tab:resolution}
\end{table}

\subsection{Ablation Study} 

All ablation studies are based on HunyuanVideo with videos generated under the configuration of 1280$\times$720 resolution with a frame number of 81 (totally 75.6K tokens). Due to space limitations, we analyze the results from the following three perspectives only. The remaining component ablation studies can be found in Appendix A.4.

% \noindent\textbf{Token Reuse}

% \noindent\textbf{Key Module Efficiency Ablation.} Three different groups are designed for comparison: 1. the complete \sysname (Ours) including "layer-wise token analysis + threshold-aware K-means (with cluster center reuse) + upper-bound approximate Top-K selection"; 2. Ablation 1, without "cluster center reuse"; 3. Ablation 2, without "Top-K selection" where all clusters participate in attention computation. These results are presented in Table 2, which shows that the dynamic clustering approach (Ours) exhibits significantly lower latency compared to the other two ablation groups. \jc{Ablation 2 does not seem reasonable, coonsider to re-design this experiment}

% \begin{table}[h]
%     \centering
%     \setlength{\tabcolsep}{8pt}
%     \begin{tabularx}{\linewidth}{
%         >{\centering\arraybackslash}X |
%         >{\centering\arraybackslash}X |
%         >{\centering\arraybackslash}X |
%         >{\centering\arraybackslash}X
%     }
%     \toprule
%          Method&   AdaCluster&  Ablation 1& Ablation 2\\\midrule
%          Latency&  \textbf{1104s}&  1247s&  4142s\\ \bottomrule
%     \end{tabularx}
%     \caption{Impact of Key Module Removal on KV-Clus Latency}
%     \label{tab:placeholder}
% \end{table}

\noindent\textbf{Cluster count strategy.} To further validate the impact of assigning an appropriate number of clusters to each layer, we compare \sysname with another strategy, `AvgClus', which assigns the same number of clusters to every layer while maintaining the same average number of clusters (specifically, 412 for the selected configuration) to ensure a fair comparison. Shown in Table~\ref{tab:adaptive-cluster-num}, `AdaClus' consistently outperforms `AvgClus' in terms of accuracy, demonstrating that adaptively determining the cluster count for each layer can effectively enhance generation quality.
\begin{table}[h]
    \centering
    \setlength{\tabcolsep}{3pt}  
    \caption{Accuracy of AdaCluster and AverageCluster.}
    \begin{tabularx}{\linewidth}{
        >{\hsize=0.8\hsize\centering\arraybackslash}X |
        >{\hsize=0.55\hsize\centering\arraybackslash}X |
        >{\hsize=0.55\hsize\centering\arraybackslash}X |
        >{\hsize=0.55\hsize\centering\arraybackslash}X |
        >{\hsize=0.85\hsize\centering\arraybackslash}X 
    }
    \toprule
         &  PSNR↑&  SSIM↑&  LPIPS↓&  ImgeQual↑ \\\midrule
         AdaClus&  30.580&  0.835&  0.203&  65.11\%\\
         AvgClus&  29.007&  0.724&  0.378&  64.79\%\\ \bottomrule
    \end{tabularx}
    \label{tab:adaptive-cluster-num}
\end{table}

% The experimental results show that the dynamic clustering strategy achieves a higher generation accuracy than the average clustering strategy and the maximum clustering strategy. Regarding the lower accuracy of the maximum clustering strategy, we attribute this phenomenon to the fact that appropriately reducing the number of clusters can yield better performance under specific token distribution patterns. For instance, if tokens in a certain layer are concentrated in a single region, forcing all clusters to participate in attention computation (i.e., not reducing the number of clusters) disrupts the aggregation of tokens close to the cluster center. This unnecessary retention of redundant clusters introduces noise into the attention calculation process, ultimately leading to lower generation accuracy compared to the dynamic clustering strategy that adapts cluster counts to distribution characteristics.

\noindent \textbf{TensorQuest accuracy.} We evaluate the accuracy gains from using TensorQuest (Sec~\ref{sec:tensorquest}) for cluster selection, comparing against a mean-based baseline (w/o Quest). As shown in Table~\ref{tab:quest-ablation}, our method achieves higher accuracy, demonstrating that TensorQuest effectively guarantees the selection of important tokens.

\begin{table}[h]
    \centering
    \setlength{\tabcolsep}{3pt}  
    \caption{Accuracy of TensorQuest and w/o Quest.}
    \begin{tabularx}{\linewidth}{
        >{\hsize=0.8\hsize\centering\arraybackslash}X |
        >{\hsize=0.55\hsize\centering\arraybackslash}X |
        >{\hsize=0.55\hsize\centering\arraybackslash}X |
        >{\hsize=0.55\hsize\centering\arraybackslash}X |
        >{\hsize=0.85\hsize\centering\arraybackslash}X 
    }
\toprule
          & PSNR↑  & SSIM↑ & LPIPS↓ & ImgeQual↑\\ \midrule
TensorQuest  & 30.580 & 0.835 & 0.203 & 65.11\%\\
w/o Quest & 28.941 & 0.687 & 0.410 & 64.06\%\\ \bottomrule
\end{tabularx}
\label{tab:quest-ablation}
\end{table}

\begin{figure}[t!]
    \centering
    \includegraphics[width=0.85\linewidth]{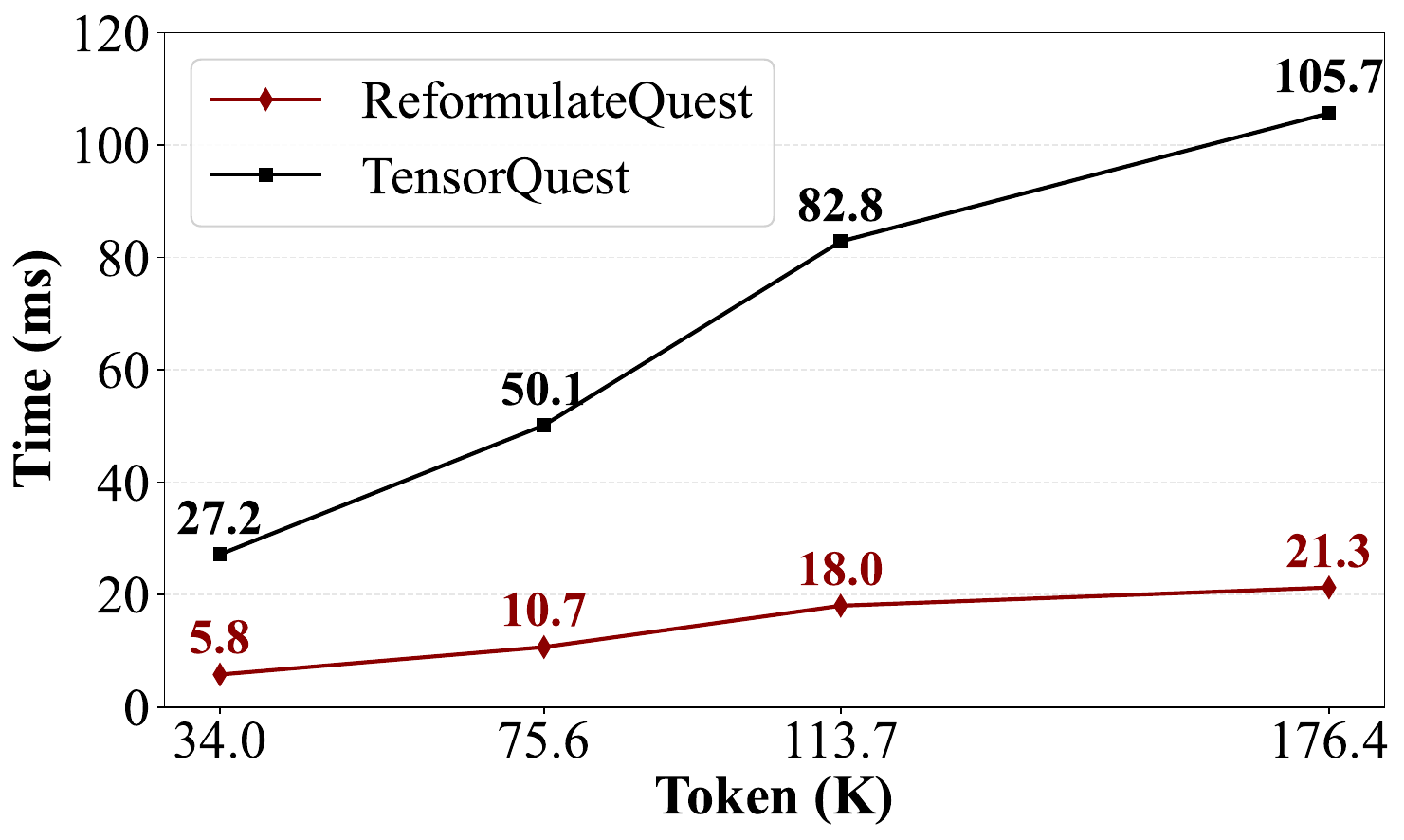}
    \caption{Top-K selection time under different input tokens.}
    \label{fig:quest-efficiency}
\end{figure}

\noindent\textbf{TensorQuest reformulation benefits.} Beyond the accuracy ablation study, we further evaluate the performance improvements achieved by the TensorQuest algorithm. As shown in Figure~\ref{fig:quest-efficiency}, the execution times under varying input token lengths are reported. For the longest input sequence, our proposed design attains a speedup of up to 5$\times$. This performance gain stems from the efficient utilization of GPU tensor cores, which are specialized for high-throughput matrix operations, whereas the original Quest computation relies solely on the less efficient CUDA cores.

\section{Related Work}

\if0 %ZW
\para{Sparse attention methods.}
%ZW \subsection{Sparse Attention Methods}
Sparse attention reduces the quadratic complexity of vanilla attention by restricting or selecting interaction patterns. Fixed-pattern methods such as Longformer~\cite{beltagy2020longformer} and BigBird~\cite{zaheer2020bigbird} adopt sliding-window and mixed global/random patterns to achieve near-linear complexity. Content-aware methods dynamically select tokens or groups: Reformer~\cite{kitaev2020reformer} leverages LSH to group similar queries/keys; Performer~\cite{choromanski2021performer} uses random feature maps for kernelized softmax approximation; Sparse Transformer~\cite{child2019sparse} designs strided/fixed sparse layouts; Routing Transformer~\cite{roy2021routing} routes queries to relevant key clusters.
Compared to these approaches, our method introduces layer-wise adaptive clustering guided by compressibility analysis, threshold-aware iterative clustering to avoid premature boundary assignment, and a Tensor Core–friendly reformulation for efficient upper-bound scoring. Moreover, we explicitly preserve hard-to-compress layers with full attention, maintaining quality while retaining most of the efficiency gains.

\fi

\para{Diffusion model quantization.}
Quantization has been widely studied to accelerate diffusion models and reduce memory usage. For example, PTQD~\cite{he2023ptqd} and PQD~\cite{ye2024pqd} investigate post‑training quantization for diffusion pipelines, while Q‑DM~\cite{li2023qdm} and Q‑Diffusion~\cite{lee2023accuquant} target low‑bit quantization. On the other hand, low‑rank quantization methods such as SVDQuant~\cite{li2024svdquant} and IntLoRA~\cite{guo2025intlora} exploit a low‑rank adapter in conjunction with quantization to further improve efficiency. These methods are orthogonal to ours: they lower numerical precision, while we reduce attention workload through token clustering and selection.
% They can be combined for cumulative gains.

\para{Training-based sparse attention.} 
VSA~\cite{zhang2025vsa} learns dynamic sparsity via two-stage attention: coarse tiling selects critical tokens, followed by fine token-level attention within selected tiles.
DSV~\cite{tan2025dsv} uses low-rank attention predictors with reduced heads, trained separately without full end-to-end integration.
RadialAttention~\cite{radial} extends generation length in pre-trained video DiT models via efficient LoRA fine-tuning.
However, these approaches generally incur substantial re-training or fine-tuning costs.
\if0 %ZW
\para{System-level optimizations.}
%ZW \subsection{System-Level Optimizations}
System-level attention optimizations such as FlashAttention~\cite{dao2022flashattention,dao2023flashattention2} fuse operators and reorder computation to reduce memory traffic and improve GPU utilization. These are complementary to our algorithmic sparsification: our Tensor Core–optimized cluster selection reduces the number of tokens to process, and fused kernels can further speed up the computation on the surviving tokens, yielding multiplicative benefits.
\fi
\section{Conclusion}

% We accelerate video diffusion transformers by exploiting token clustering for sparse attention. We reveal that token distributions exhibit varying compressibility across different layers, and we propose AdaCluster, a training-free method that adaptively clusters tokens layer-wise to reduce attention computation, including layer-wise compressibility analysis, distribution-aware adaptive clustering, and upper-bound based Top-K cluster selection. On representative open video diffusion transformers (CogVideoX-2B, HunyuanVideo, and Wan-2.1), AdaCluster demonstrates prominent end-to-end speedup (1.67$\times$--4.31$\times$)  with negligible performance degradation. %4.31，hunyuan最大尺寸
We introduced \sysname, a training‑free adaptive query–key clustering framework for sparse attention in DiTs. By tailoring clustering strategies to the heterogeneous distributions of query and key vectors, and by integrating efficient critical token selection via TensorQuest, \sysname achieves substantial improvement in generation speed while preserving high‑fidelity video quality.
\section*{Acknowledgements}
We thank the anonymous reviewers for their insightful comments. This work was supported by the National Natural Science Foundation of China under Grant No. 62572454. This work is done with the sponsorship of TeleAI. The corresponding author is Youhui Bai.
%\clearpage
%\input{rebuttal}
%\clearpage

% \input{sec/3_finalcopy}
{
    \small
    \bibliographystyle{ieeenat_fullname}
    \bibliography{main}
}


\onecolumn
\appendix

\section{Appendix}

\subsection{Compactness with Different Request}
\begin{figure*}[htbp]
    \centering
    \includegraphics[width=0.32\textwidth]{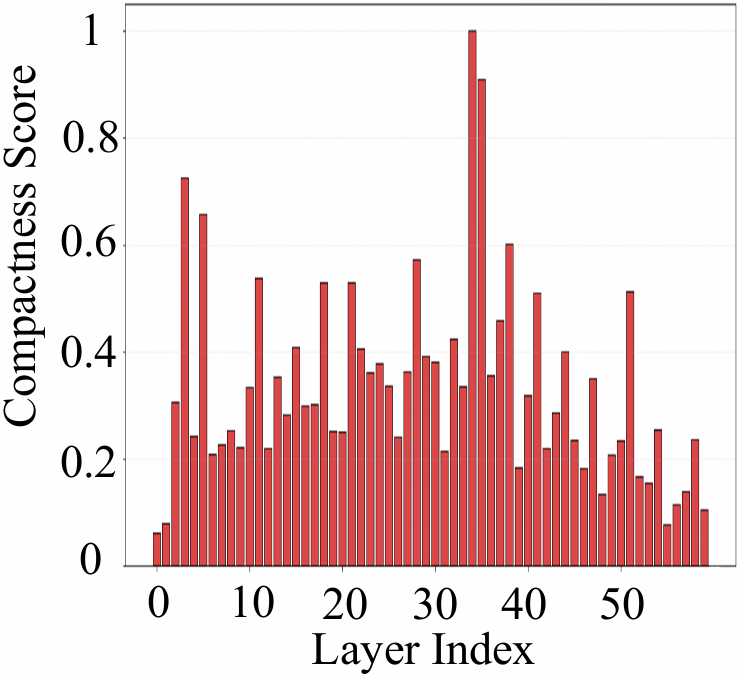}
    \hfill
    \includegraphics[width=0.32\textwidth]{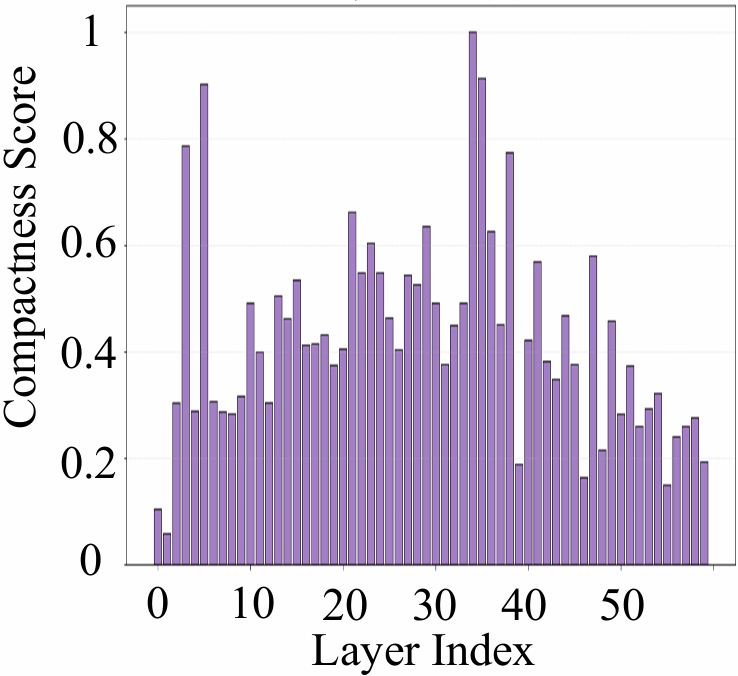}
    \hfill
    \includegraphics[width=0.32\textwidth]{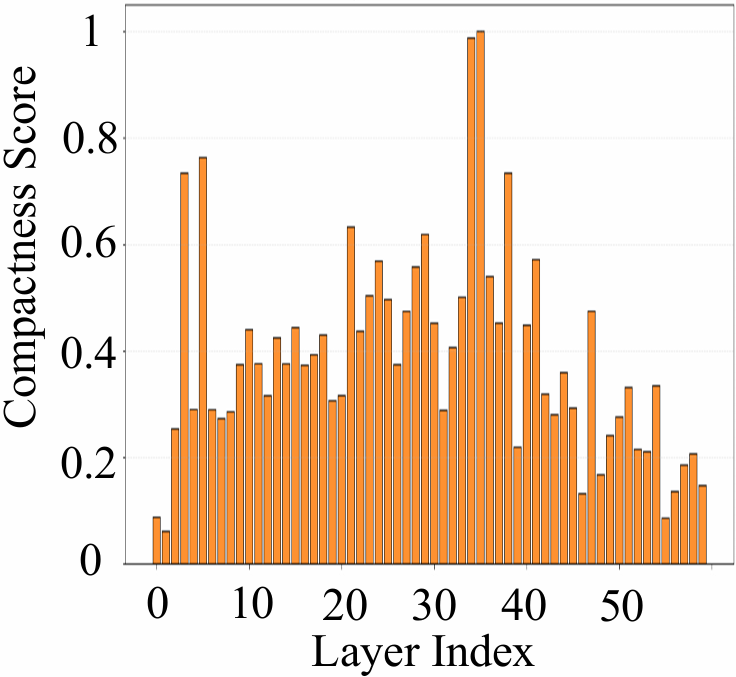}
    \caption{Compactness of each model layer under different inputs. (a) Subfigure 1 (red): Prompt is "In the large cage, two puppies were wagging their tails at each other." (b) Subfigure 2 (purple): Prompt is "A flock of bats flies over the village, captured in medium long shot." (c) Subfigure 3 (orange): Prompt is "Above the sea, a school of silver flying fish leaped out of the water."}
    \label{fig:compact-prompt}
\end{figure*}

To verify the effectiveness of the method proposed in Section 3.2.1, we selected the first three prompts from PenguinVideoBenchmark as inputs and tested the compactness of each layer of the HunyuanVideo model under different prompts during the inference of 720p videos. The results are presented in Figure \ref{fig:compact-prompt}. We can observe that for different inputs, the compactness of the model layers exhibits similar trends under the same configuration, which validates the effectiveness of the method proposed in Section 3.2.1.

\clearpage
\subsection{Theoretical and Empirical Justification of Query Normalization}
\textbf{Theoretical justification.}
Let $q \in \mathbb{R}^d$ be a query vector, $k \in \mathbb{R}^d$ a key vector, and $s = q^\top k$ the attention score. After normalizing $q$ to $\hat{q} = q / \|q\|_2$, the score becomes $\hat{s} = \hat{q}^\top k = s / \|q\|_2$. Since $\|q\|_2 > 0$, for any pair of keys $k_a, k_b$ we have $q^\top k_a > q^\top k_b \iff \hat{q}^\top k_a > \hat{q}^\top k_b$. Thus the relative ordering of scores—and hence the Top-$K$ selection—is invariant under query normalization, which justifies the statement in Sec.~3.1 that “the relative magnitude of query–key scores is independent of the query vector length.”

\begin{figure*}[htbp]
    \centering
    \includegraphics[width=\textwidth]{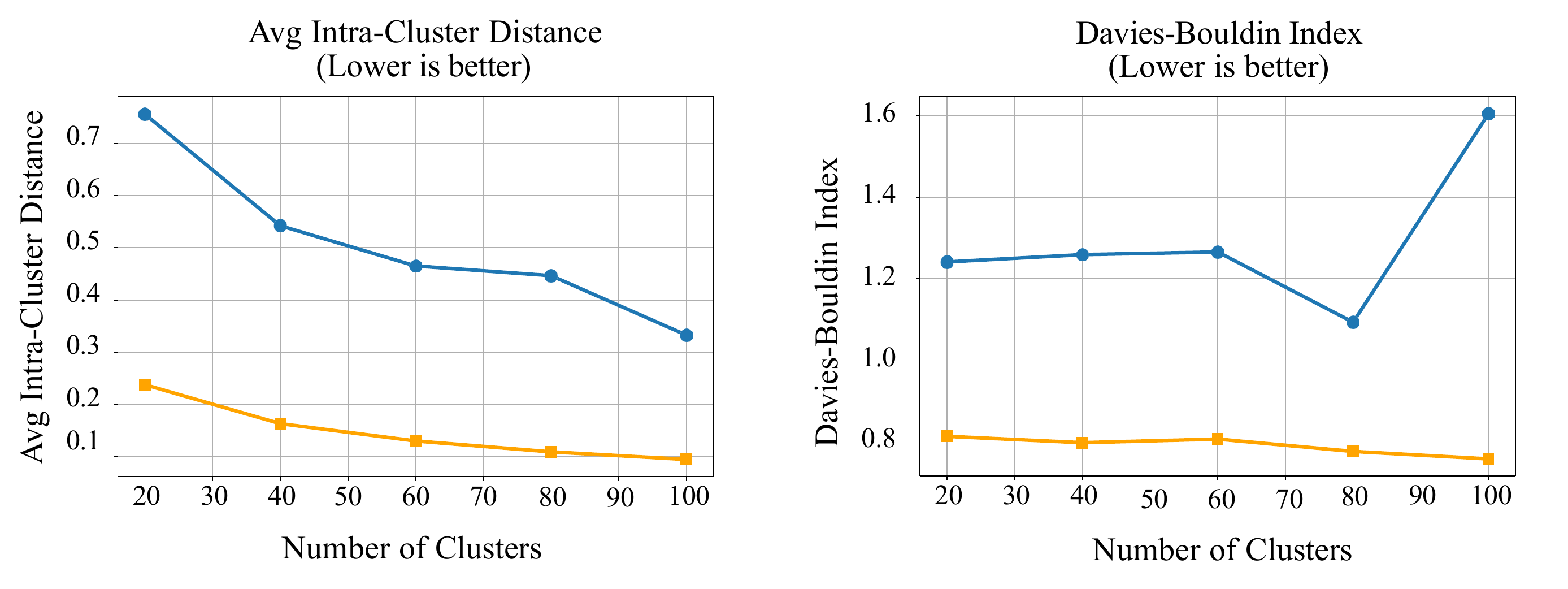}
    \caption{Left: average intra-cluster distance (lower is better). Right: Davies–Bouldin index (lower is better). Both curves compare clustering unnormalized queries (“Original”) versus clustering normalized queries (“Normalized”) under identical numbers of clusters.}
    \label{fig:qclus}
\end{figure*}

\textbf{Empirical evidence (Figure~\ref{fig:qclus}).}
We compare clustering unnormalized queries (“Original”) versus clustering normalized queries (“Normalized,” our method). Both approaches are evaluated in the same normalized space. The left panel of Figure~\ref{fig:qclus} plots the average intra-cluster distance as a function of the number of clusters $K$, showing that normalized queries yield consistently tighter clusters for all $K$. The right panel reports the Davies–Bouldin (DB) index, whose definition for $K$ clusters is
\[
DB = \frac{1}{K} \sum_{i=1}^K \max_{j \ne i} \frac{S_i + S_j}{M_{ij}},
\]
where $S_i$ is the average intra-cluster distance of cluster $i$ and $M_{ij}$ denotes the distance between centroids $i$ and $j$. A smaller DB index indicates more compact clusters with better inter-cluster separation; again, normalization clearly improves the result across all $K$.

In our deployment we fix the normalized approach to $K=65$ clusters (the setting used in Sec.~3.1) and ask how many clusters the unnormalized approach needs to match the same intra-cluster distance in the normalized space. On average the unnormalized clustering requires over 235 clusters, corresponding to an effective compression ratio of about $3.6\times$. These observations confirm that normalizing queries before clustering not only preserves Top-$K$ correctness, but also makes the queries substantially easier to cluster—yielding higher compression ratios and tighter clusters as claimed in Sec.~3.1.

To further assess the practical effect of query normalization on reconstruction quality, we conduct an ablation study on the Hunyuan dataset. The following table compares key metrics with and without query normalization (all other components remain identical):

\begin{table}[ht]
\centering
\caption{Impact of query normalization on reconstruction quality (Hunyuan)}
\begin{tabular}{lcc}
\toprule
Metric & Without Normalization & With Normalization (Ours) \\
\midrule
PSNR $\uparrow$   & 29.56 & \textbf{30.58} \\
SSIM  $\uparrow$   & 0.763 & \textbf{0.835} \\
LPIPS $\downarrow$ & 0.317 & \textbf{0.203} \\
\bottomrule
\end{tabular}
\end{table}

\clearpage
\subsection{Number of Clusters for k Under Different Videos Configuration}
\begin{figure*}[htbp]
    \centering
    \captionsetup[subfigure]{labelformat=empty}
    
    \begin{subfigure}{0.7\textwidth}
        \centering
        \includegraphics[width=\linewidth]{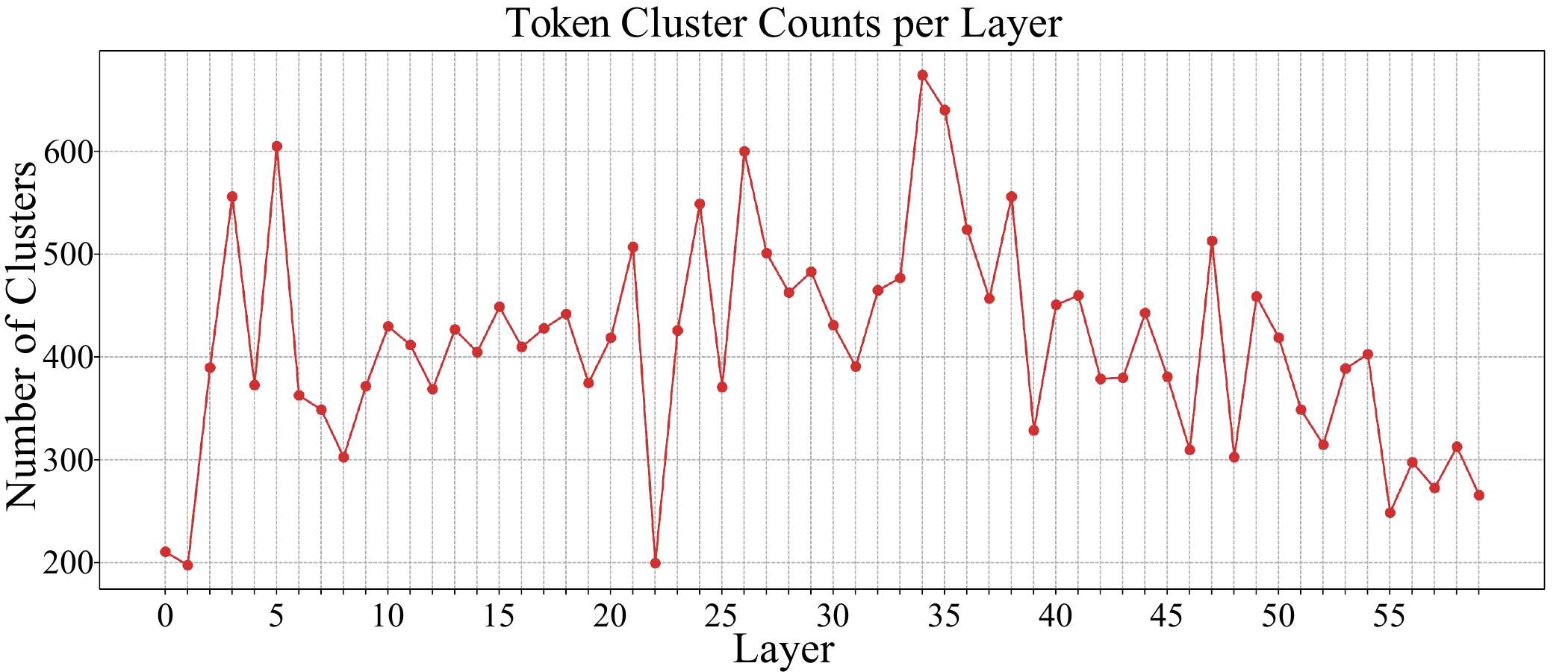}
        \caption{Configuration: 1920$\times$1120} 
        \label{fig:sub1}
    \end{subfigure}
    \vspace{5pt}
    
    \begin{subfigure}{0.7\textwidth}
        \centering
        \includegraphics[width=\linewidth]{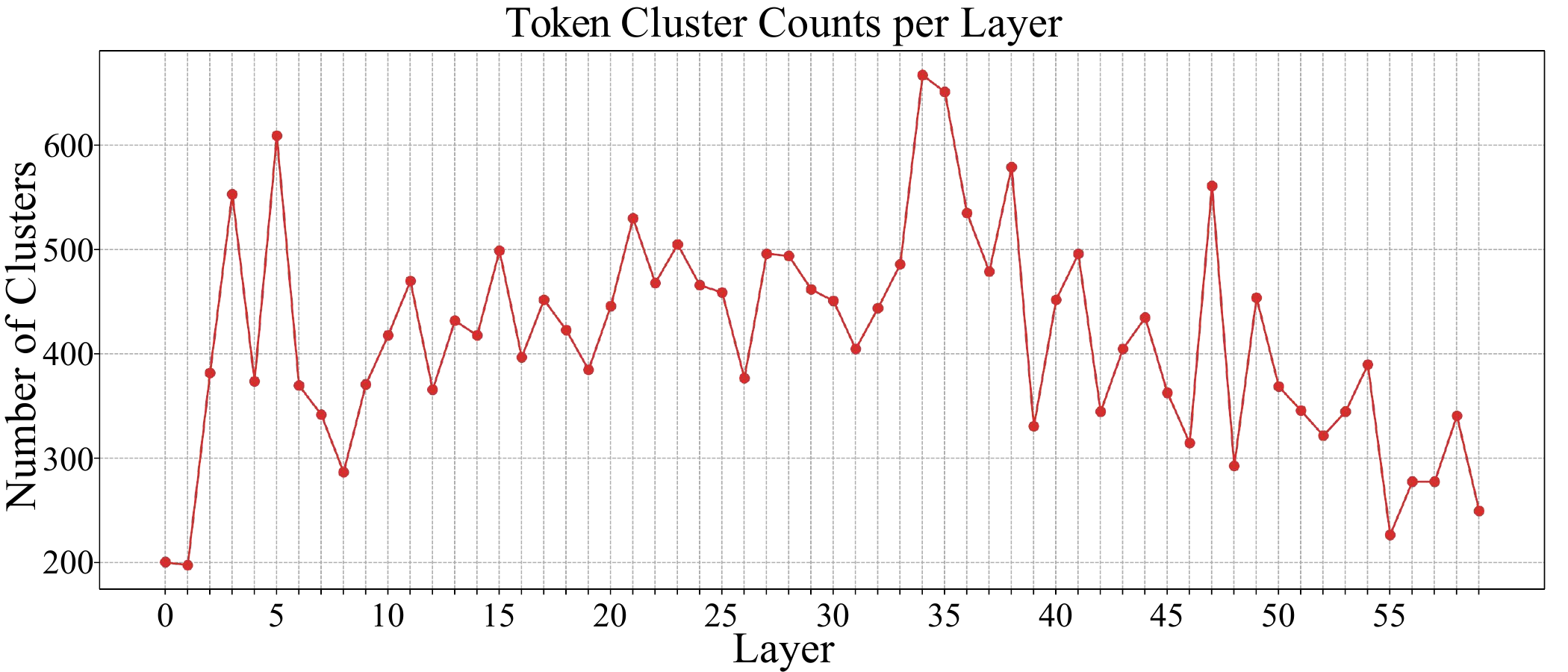}
        \caption{Configuration: 1280$\times$720}
        \label{fig:sub2}
    \end{subfigure}
    \vspace{5pt}
    
    \begin{subfigure}{0.7\textwidth}
        \centering
        \includegraphics[width=\linewidth]{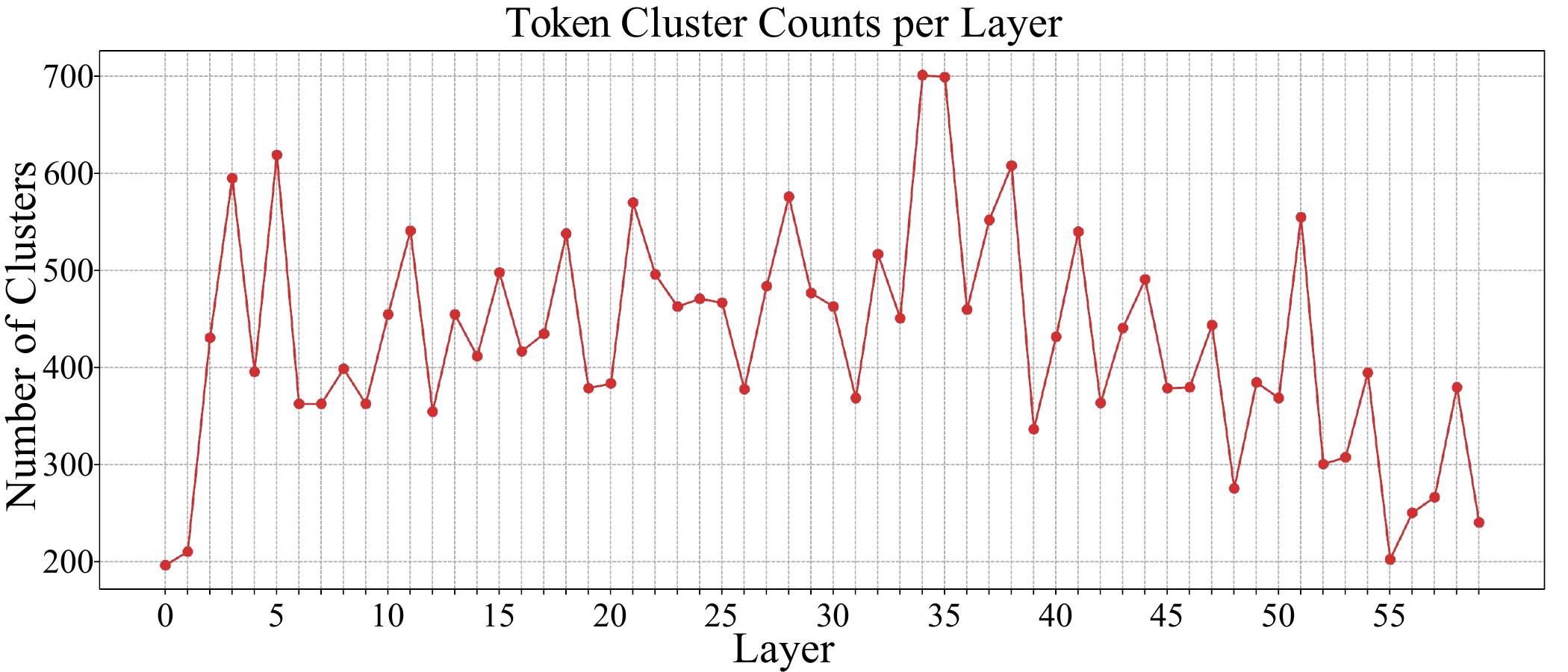}
        \caption{Configuration: 720$\times$480}
        \label{fig:sub3}
    \end{subfigure}
    
    \caption{Number of clusters for k under different videos configuration}
    \label{fig:cluster-config}
\end{figure*}

On the Hunyuan model, we measured the number of K clusters for each layer under different configurations, as shown in Figure \ref{fig:cluster-config}. It can be observed that the trends of the number of K clusters across model layers are roughly the same under different configurations, though there are still slight differences. The dynamic clustering approach can better balance the accuracy and speed of video generation.

\clearpage

\subsection{Component Ablation}

To better understand the contribution of each component in AdaCluster, 
we summarize the ablation results of three key components in Table~\ref{tab:component_ablation}: 
(1) adaptive key clustering, (2) TensorQuest-based cluster selection, and 
(3) query normalization.

Starting from a baseline with mean-based clustering and without TensorQuest, 
introducing TensorQuest improves token selection quality and reconstruction accuracy.
Replacing mean-based clustering with our adaptive clustering strategy further boosts performance,
demonstrating the benefit of adaptive key grouping.
Finally, applying query normalization stabilizes similarity estimation and provides the largest improvement across all metrics.

\begin{table}[h]
\centering
\small
\begin{tabular}{lccc|cccc}
\toprule
\textbf{Configuration} 
& \textbf{PSNR} $\uparrow$ & \textbf{SSIM} $\uparrow$ & \textbf{LPIPS} $\downarrow$ & \textbf{ImageQual} $\uparrow$ \\
\midrule
AvgCluster + w/o Quest + w/o Norm & 28.94 & 0.687 & 0.410 & 64.06\% \\
AvgCluster + TensorQuest + w/o Norm  & 29.01 & 0.724 & 0.378 & 64.79\% \\
AdaCluster + TensorQuest + w/o Norm & 29.56 & 0.763 & 0.317 & 65.03\% \\
\textbf{AdaCluster + TensorQuest + Norm (Ours)}& \textbf{30.58} & \textbf{0.835} & \textbf{0.203} & \textbf{65.11\%} \\
\bottomrule
\end{tabular}
\caption{
Component analysis of AdaCluster. 
}
\label{tab:component_ablation}
\end{table}

\subsection{Hyperparameter Sensitivity Analysis}

To provide a thorough understanding of AdaCluster's robustness, we conduct a grid search over five key hyperparameters on the Hunyuan reconstruction task. The table below reports the reconstruction quality (PSNR, SSIM, LPIPS) and inference speedup under different configurations. The setting used in the main paper is highlighted in \textbf{bold}.

\begin{table}[ht]
\centering
\caption{Hyperparameter sensitivity analysis on Hunyuan. The configuration used in the main experiments is highlighted in \textbf{bold}.}
\small
\begin{tabular}{ccccccccc}
\toprule
Sparsity & KV Thresh & Q Clus &Initial K&TopK & PSNR$\uparrow$ & SSIM$\uparrow$ & LPIPS$\downarrow$ & Speedup \\
\midrule
74.8\% & 5.5 & 30  & 50  & 64 & 30.21 & 0.820 & 0.220 & 1.72$\times$ \\
77.5\% & 3.5 & 65  & 100 & 64 & 31.88 & 0.850 & 0.180 & 1.69$\times$ \\
\textbf{76.4\%} & \textbf{5.5} & \textbf{65} & \textbf{100} & \textbf{64} & \textbf{30.58} & \textbf{0.835} & \textbf{0.203} & \textbf{1.68$\times$} \\
60.1\% & 8.0 & 65  & 100 & 64 & 30.58 & 0.825 & 0.233 & 1.55$\times$ \\
83.2\% & 5.5 & 65  & 200 & 32 & 28.82 & 0.635 & 0.395 & 1.90$\times$ \\
\bottomrule
\end{tabular}
\end{table}

As shown, the selected configuration (76.4\% sparsity, KV threshold 5.5, 65 query clusters, initial Top-100, codebook size 64) achieves a strong balance between reconstruction fidelity and inference speedup. Higher sparsity or larger initial TopK tends to yield greater acceleration at the cost of quality degradation, while more aggressive KV thresholding or fewer query clusters can improve quality in some regimes but reduces overall speedup.
\clearpage

\subsection{Additional Results on H100 GPUs}

\subsubsection{Setup}

To further validate the scalability of \textbf{AdaCluster} on larger models and longer sequence lengths, we conduct additional experiments on NVIDIA H100 GPUs (80GB HBM). We evaluate two representative high-capacity Diffusion Transformer  based Text-to-Video models: Wan2.1-T2V-14B and HunyuanVideo.

\textbf{Models and Configurations.}  
For \textbf{Wan2.1-T2V-14B}, we generate 5-second videos at $832 \times 480$ resolution using 50 inference steps.  
For \textbf{HunyuanVideo}, we generate videos at $1280 \times 720$ resolution using 30 inference steps (consistent with the main experiments).  

All other hyperparameters remain identical to those described in Section~4.1: $k_{\max}$ is set such that the top 15\% of layers use full attention, $\tau = 1.5\times$ the average token-to-cluster-center distance from the first inference step, the number of query clusters is fixed at 65, and key clusters are dynamically adjusted. Cluster centers from the previous timestep are reused for subsequent steps. We integrate the same Triton-based clustering kernel and FlashAttention fallback for skipped layers.

\textbf{Datasets and Metrics.}  
We use the same prompt set from PenguinVideoBenchmark as in the main experiments. Efficiency is measured by end-to-end generation speedup (relative to the FlashAttention baseline) and total attention computation in PFLOPs.

\subsubsection{Results and Analysis}
Table~\ref{tab:h100_results} summarizes the efficiency and quality results on H100.

\begin{table}[htbp]
\centering
\caption{Efficiency and quality results of AdaCluster and baselines on H100.}
\label{tab:h100_results}
\begin{tabular}{lccccc}
\toprule
Model & Method & Speedup$\uparrow$ & Attention (PFLOPs)$\downarrow$  \\
\midrule
Wan2.1-14B & SVG2 & 1.61$\times$ & 427.43  \\
Wan2.1-14B & AdaCluster & \textbf{1.81$\times$} &  \textbf{404.94}  \\
HunyuanVideo & SVG2 & 1.58$\times$ & 335.36\\
HunyuanVideo & AdaCluster & \textbf{1.67$\times$}  & \textbf{322.21} \\
\bottomrule
\end{tabular}
\end{table}
% \textcolor{white}{.}\newline

The results in Table~\ref{tab:h100_results} demonstrate that AdaCluster consistently outperforms SVG2 across both models. On Wan2.1-14B, AdaCluster achieves a speedup of 1.81$\times$ (vs. 1.61$\times$ for SVG2) while reducing attention PFLOPs from 427.43 to 404.94. Similarly, on HunyuanVideo, AdaCluster attains a speedup of 1.67$\times$ (vs. 1.58$\times$) and lowers attention PFLOPs to 322.21 from 335.36. These improvements confirm the effectiveness of adaptively allocating cluster numbers per layer in enhancing both efficiency and generation quality.
% 如果你有具体的 attention density 数值，可以在这里添加或新建一个小节
% \subsection{Attention Density Analysis}
% AdaCluster achieves an average attention density of XX\% versus YY\% for SVG2 on HunyuanVideo (see Figure~\ref{fig:attn_density_h100}).

% 图表建议（请自行插入实际图片）
% \begin{figure}[htbp]
%   \centering
%   \includegraphics[width=0.9\linewidth]{figures/h100_speedup_flops.pdf}
%   \caption{Speedup and attention FLOPs comparison on H100.}
%   \label{fig:h100_speedup}
% \end{figure}

% \begin{figure}[htbp]
%   \centering
%   \includegraphics[width=0.9\linewidth]{figures/attn_density_h100.pdf}
%   \caption{Attention density distribution across layers on H100 (AdaCluster vs. SVG2).}
%   \label{fig:attn_density_h100}
% \end{figure}
\clearpage
\subsection{Video Results with Different Methods}
 We selected HunyuanVideo as the baseline and compared the performance of different methods across various prompts, where all videos were produced on a single NVIDIA A40 GPU with configuration under 1280$\times$720, 81 frames. The results demonstrate that AdaCluster achieves satisfactory accuracy, with high similarity to full attention. SVG2 performs reasonably well in most scenarios, but static clustering often encounters issues, particularly evident in the handling of video details. SpargeAttn, however, delivers poor performance due to neglecting embedding similarity.

\begin{figure}[htbp]
    \centering
    \includegraphics[width=1\columnwidth]{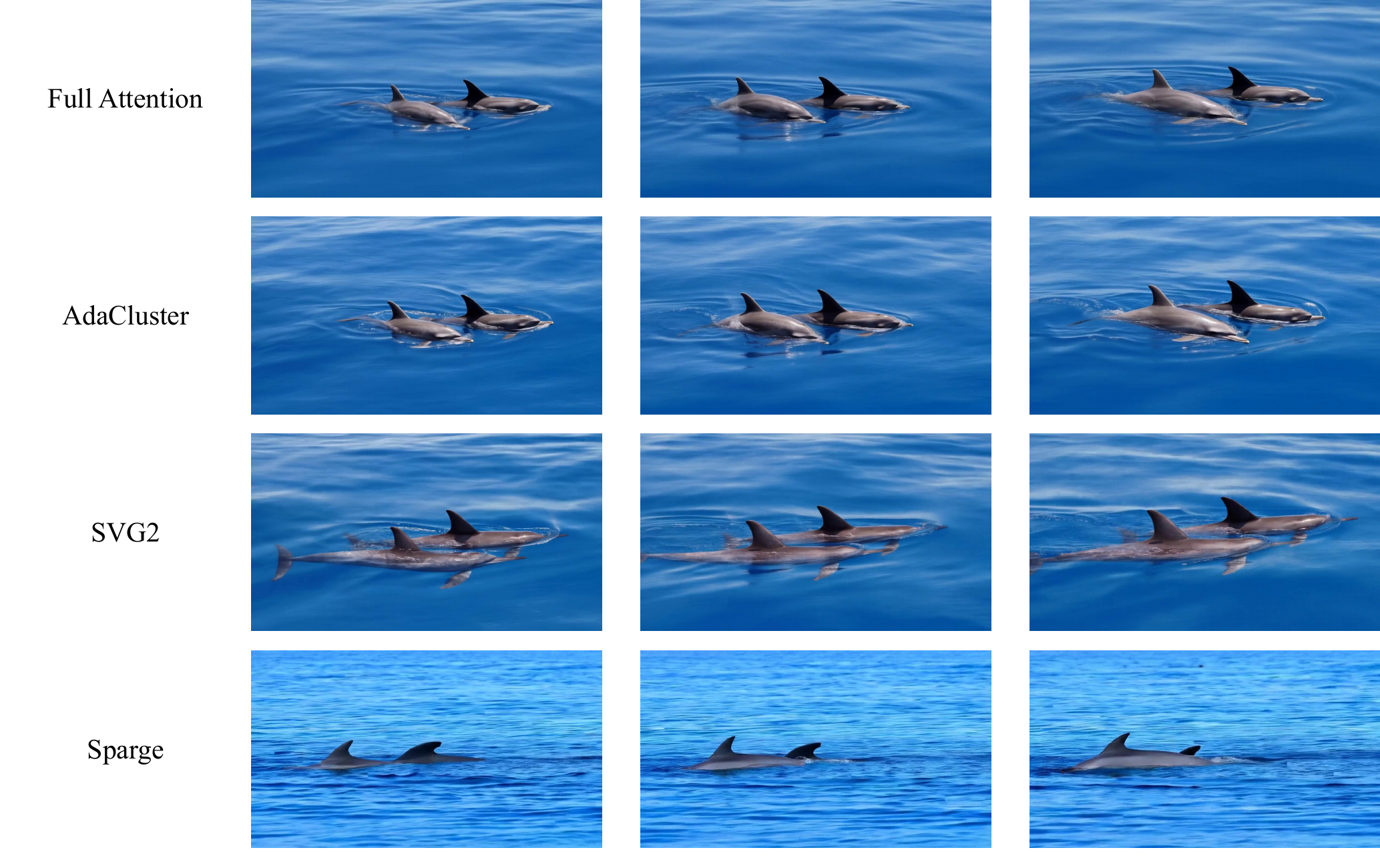}
    \caption{Prompt: Two dolphins are swimming in the blue sea.}
    \label{fig:hunyuan-screenshot}
\end{figure}

\begin{figure}[htbp]
    \centering
    \includegraphics[width=1\columnwidth]{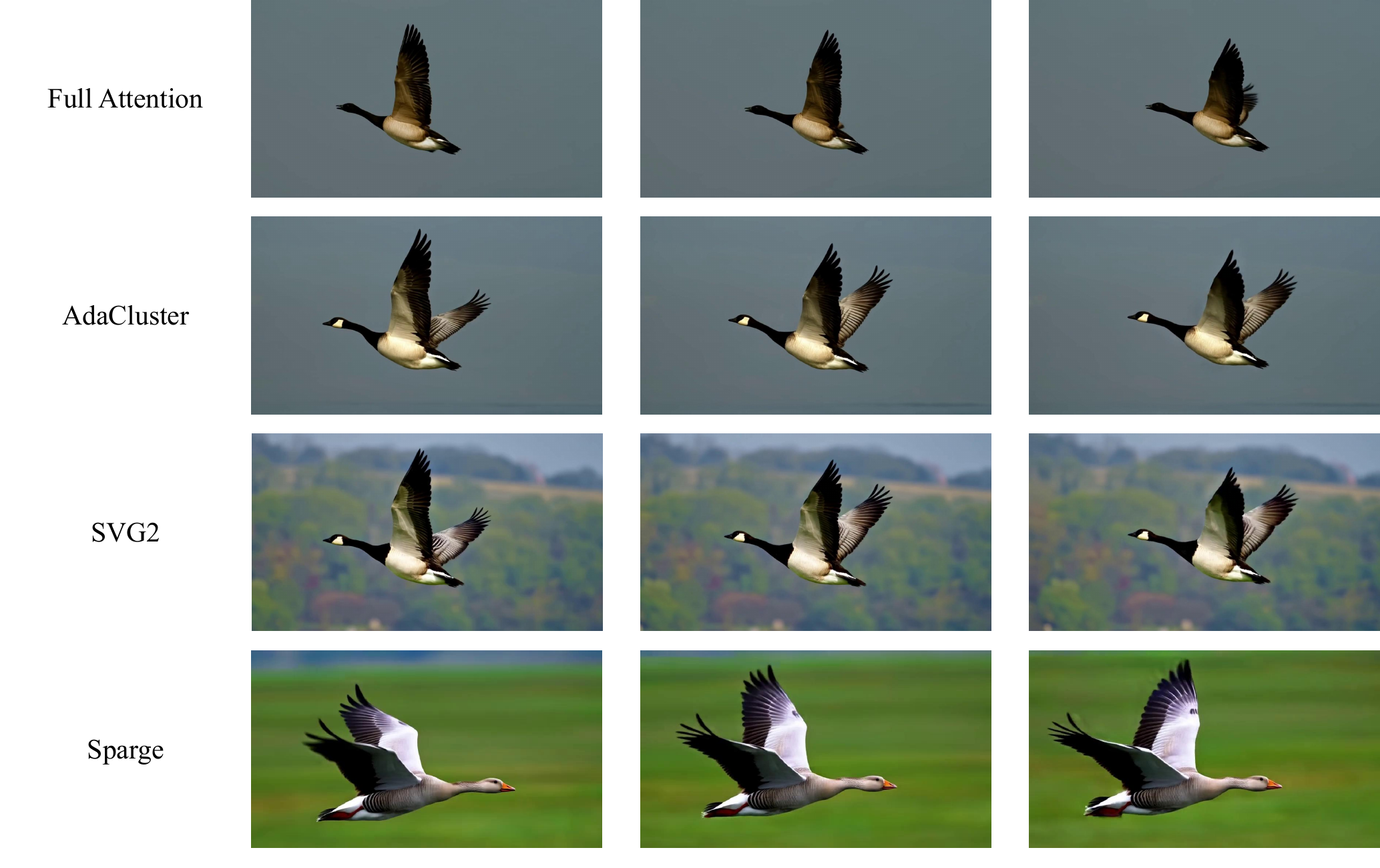}
    \caption{Prompt: A flying wild goose captured from a low-angle shot.}
    \label{fig:cog-screenshot}
\end{figure}

\begin{figure}[htbp]
    \centering
    \includegraphics[width=1\columnwidth]{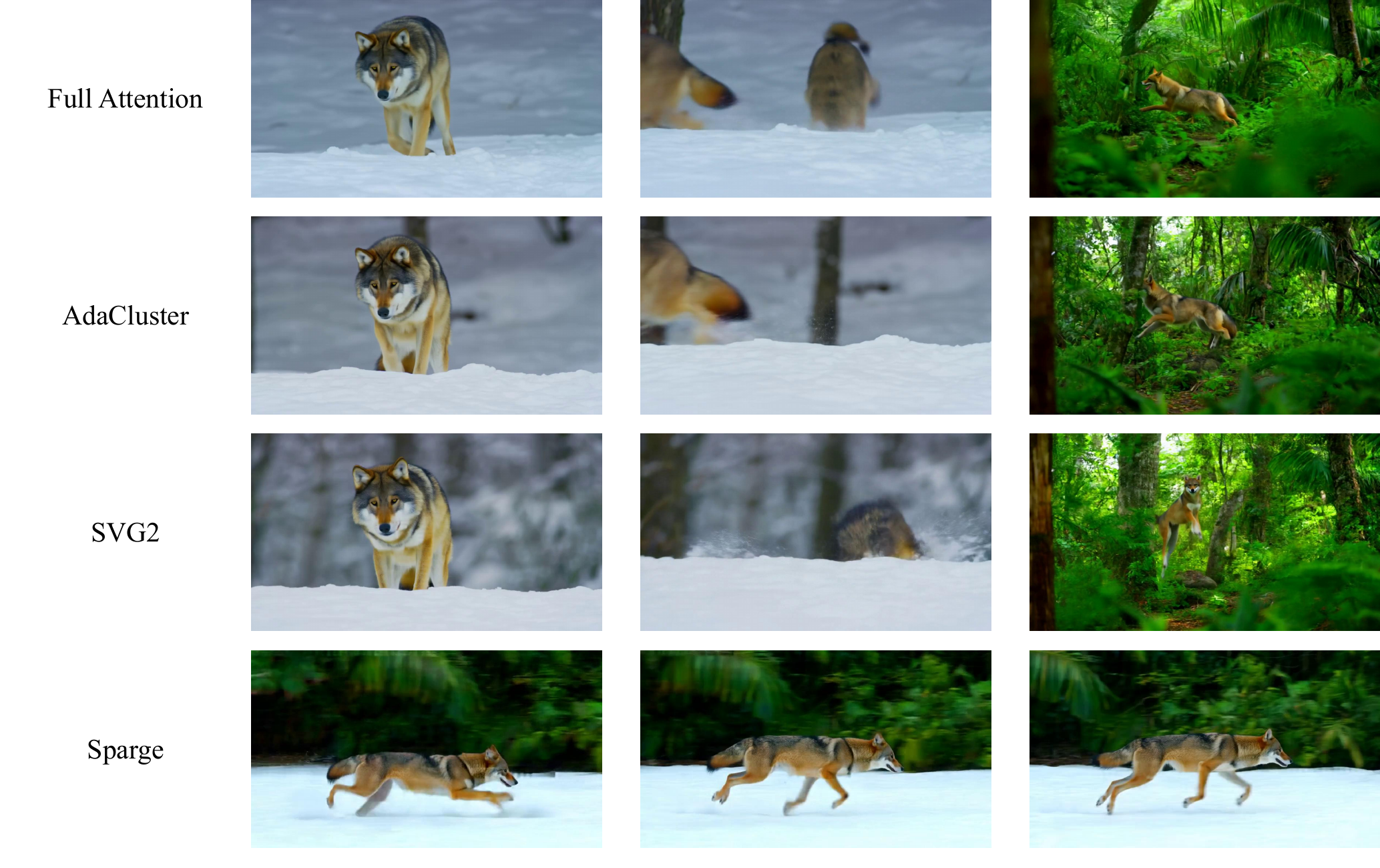}
    \caption{Prompt: Tracking shot of a wolf slowly moving through a tranquil snowy landscape, leaving footprints with each step. Suddenly, the wolf accelerates into a run and leaps. At the moment of the jump, the camera cuts to reveal the wolf landing in the midst of a lush, verdant tropical rainforest.}
    \label{fig:wan-screenshot}
\end{figure}

\clearpage
\subsection{Video Results with Different Models and Prompts} We provide video screenshots generated by AdaCluster and full attention for comparison, where all videos were produced on a single NVIDIA A40 GPU. The comparison covers three different models: HunyuanVideo, CogvideoX and Wan-2.1.

Figure~\ref{fig:cog-screenshot} shows the results on CogVideoX, Figure~\ref{fig:hunyuan-screenshot} presents the outputs of HunyuanVideo, and Figure~\ref{fig:wan-screenshot} illustrates the results obtained with Wan-2.1. As shown, AdaCluster maintains a high degree of similarity to full attention across all models, demonstrating strong generalization capability. Notably, the original output quality of CogVideoX is  inferior to that of HunyuanVideo and Wan-2.1. Nevertheless, AdaCluster consistently achieves stable and highly aligned results even on the lower-quality baseline.
\begin{figure}[htbp]
    \centering
    \includegraphics[width=1\columnwidth]{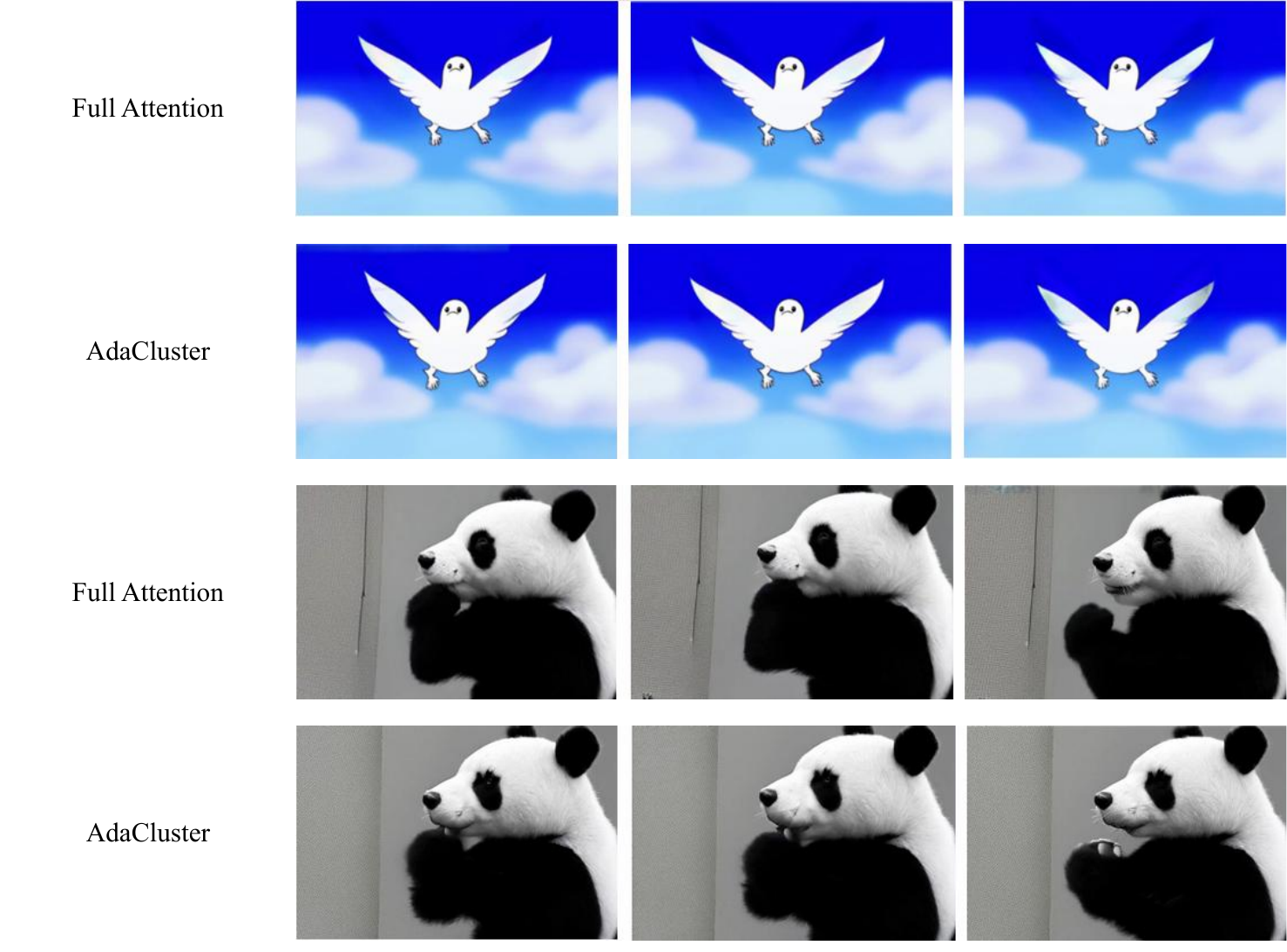}
    \caption{CogvideoX model screenshot}
    \label{fig:cog-screenshot}
\end{figure}

\begin{figure}[htbp]
    \centering
    \includegraphics[width=1\columnwidth]{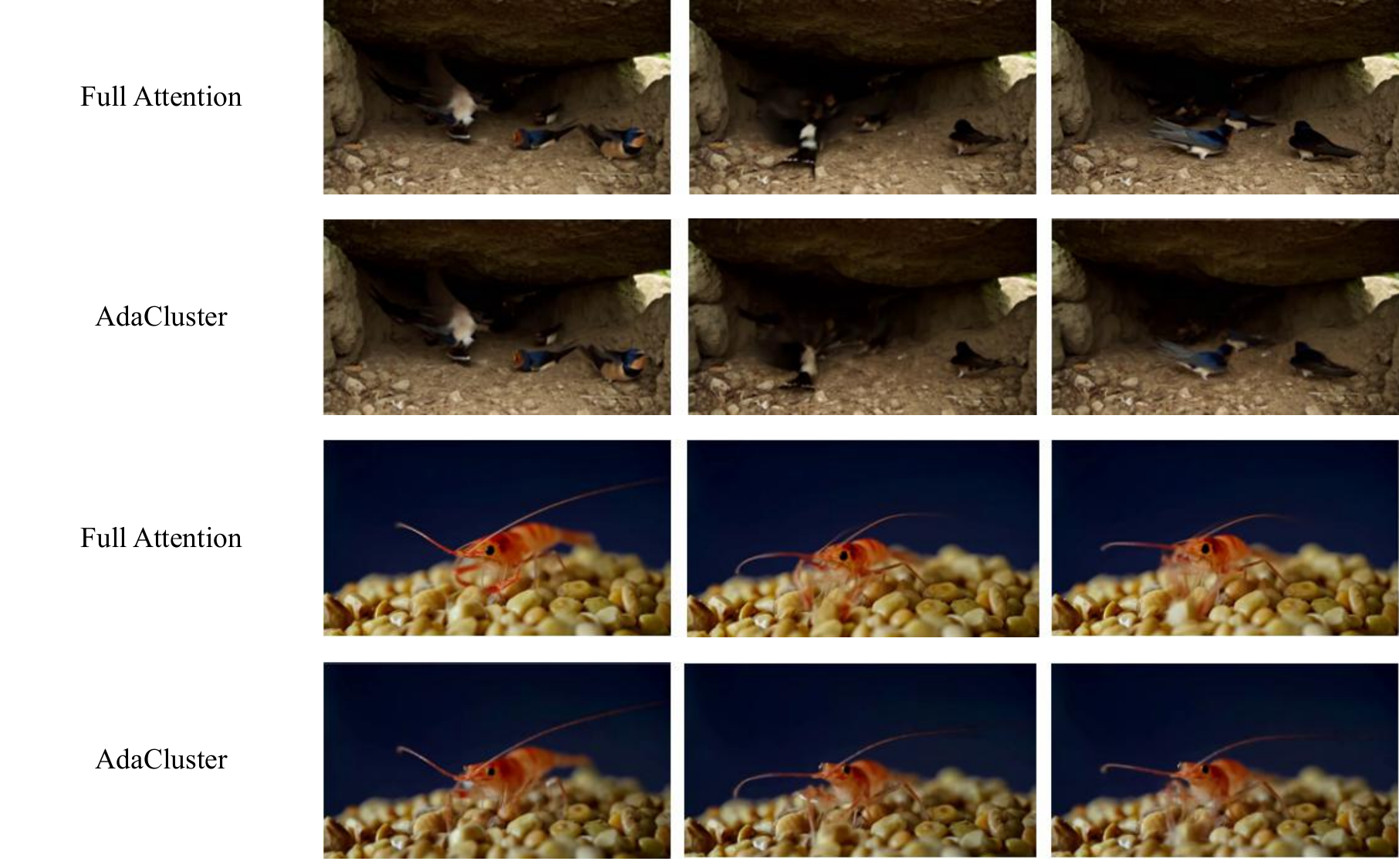}
    \caption{HunyuanVideo model screenshot}
    \label{fig:hunyuan-screenshot}
\end{figure}

\begin{figure}[htbp]
    \centering
    \includegraphics[width=1\columnwidth]{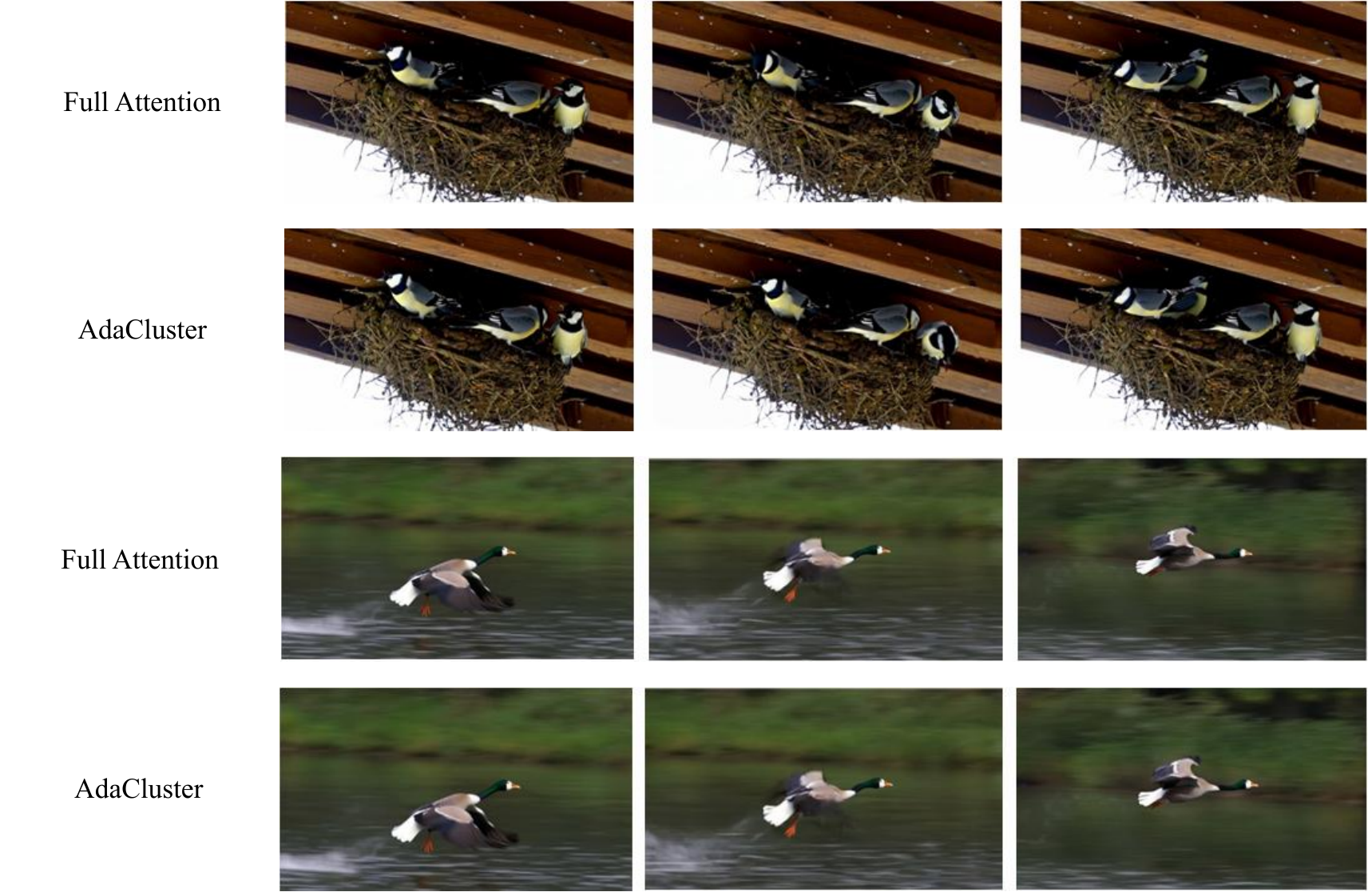}
    \caption{Wan-2.1 model screenshot}
    \label{fig:wan-screenshot}
\end{figure}

\end{document}